\documentclass[journal]{IEEEtran}

\usepackage[pdftex]{graphicx}
\usepackage[colorlinks=false]{hyperref}
\usepackage{subfigure}
\usepackage[space]{cite}
\usepackage{amsmath,amssymb,amsopn,amstext,amsfonts, amsthm}
\usepackage{amsmath,amsopn,amstext,amsfonts,mathtools}
\usepackage{booktabs}
\usepackage{xcolor}
\usepackage{threeparttable}
\usepackage{yfonts}
\usepackage{multirow}
\usepackage{bbm}
\usepackage{soul}
\usepackage[ruled,vlined,linesnumbered]{algorithm2e}
\usepackage{threeparttable}
\usepackage{pdfpages}

\hypersetup{
	hidelinks,
	colorlinks=false,
	linkcolor=black,
	citecolor=black
}

\begin{document}

\title{Voxel-SLAM: A Complete, Accurate, and Versatile LiDAR-Inertial SLAM System}

\author{Zheng Liu, Haotian Li, Chongjian Yuan, Xiyuan Liu, Jiarong Lin, Rundong Li, Chunran Zheng, Bingyang Zhou, Wenyi Liu, and Fu Zhang
\thanks{Zheng Liu, Haotian Li, Chongjian Yuan, Xiyuan Liu, Jiarong Lin, Rundong Li, Chunran Zheng, Bingyang Zhou, Wenyi Liu, and Fu Zhang are with the Department of Mechanical Engineering, The University of Hong Kong, Hong Kong (e-mail: \{u3007335, haotianl, ycj1, xliuaa, zivlin, rdli10010, zhengcr, byzhou, liuwenyi\}@connect.hku.hk; fuzhang@hku.hk)}
}


\maketitle
\begin{abstract}
In this work, we present Voxel-SLAM: a complete, accurate, and versatile LiDAR-inertial SLAM system that fully utilizes \textit{short-term}, \textit{mid-term}, \textit{long-term}, and \textit{multi-map} data associations to achieve real-time estimation and high precision mapping. The system consists of five modules: \textit{initialization}, \textit{odometry}, \textit{local mapping}, \textit{loop closure}, and \textit{global mapping}, all employing the same map representation, an adaptive voxel map. The initialization provides an accurate initial state estimation and a consistent local map for subsequent modules, enabling the system to start with a highly dynamic initial state. The odometry, exploiting the short-term data association, rapidly estimates current states and detects potential system divergence. The local mapping, exploiting the mid-term data association, employs a local LiDAR-inertial bundle adjustment (BA) to refine the states (and the local map) within a sliding window of recent LiDAR scans. The loop closure detects previously visited places in the current and all previous sessions. The global mapping refines the global map with an efficient hierarchical global BA. The loop closure and global mapping both exploit long-term and multi-map data associations. We conducted a comprehensive benchmark comparison with other state-of-the-art methods across 30 sequences from three representative scenes, including narrow indoor environments using hand-held equipment, large-scale wilderness environments with aerial robots, and urban environments on vehicle platforms. Other experiments demonstrate the robustness and efficiency of the initialization, the capacity to work in multiple sessions, and relocalization in degenerated environments. To benefit the community, we make our code publicly available\footnote{\url{https://github.com/hku-mars/Voxel-SLAM}}.
\end{abstract}

\IEEEpeerreviewmaketitle


\section{Introduction}

3D Light Detection and Ranging (LiDAR) has become a popular sensing technology in the field of autonomous mobile robots, such as autonomous driving vehicles \cite{li2020lidar} and unmanned drones \cite{kong2021avoiding, chen2023self}, due to their direct, dense, active, and accurate (DDAA) depth measurements. LiDAR simultaneous localization and mapping (SLAM) and LiDAR odometry (LO) utilize the measurements of LiDAR to provide essential state feedback and a 3D dense point cloud of surroundings for the use of subsequent processes (e.g., planning, control) on the robot. Moreover, with the development of LiDAR technology, solid-state LiDARs have been extensively investigated \cite{wang2020mems, zhang2018comparison, liu2021low} and applied in some LO \cite{lin2020loam, xu2021fast} and LiDAR SLAM systems \cite{lin2019fast, zou2024lta} due to the benefits of small size, lightweight, and affordability.


The difference between LiDAR SLAM and LiDAR odometry (LO) is that the target of LiDAR SLAM is to build a consistent map of the environment and estimate current poses within that map in real-time on a mobile robot, whereas LO systems focus on real-time localization and accumulate the map without considering refining it to mitigate drift. Following the concepts of ORB-SLAM3 \cite{campos2021orb, cadena2016past}, the superiority of LiDAR SLAM over LO is the ability to match and refine previous measurements by making full use of four types of data association:

\begin{enumerate}
	
\item Short-term data association: associating the current scan to a map and estimating the current ego-motion as efficiently as possible for the use of subsequent processes (e.g., planning, control). Most LO/LIO systems have the short-term data association \cite{xu2021fast, xu2022fast, bai2022faster, he2023point} and accumulate the current LiDAR scan into their maps without considering any further refinement of the state or map, causing cumulative drifts.

\item Mid-term data association: associating multiple latest scans to map and refining the map accordingly. This is generally achieved via a bundle adjustment (BA) technique to simultaneously refine the states and local map within a period of recent time, alleviating the cumulative drift and enhancing the robustness.

\item Long-term data association: associating all previous LiDAR scans to achieve global map consistency. Long-term data association allows to detect previously visited places to reset the drift by pose-graph optimization (PGO). Global BA could also be used to further improve the accuracy.

\item Multi-map data association: associating multiple map sessions. It uses similar methods in long-term data association to merge different map sessions, which may be caused by different collection times or re-initialization of new sessions due to LiDAR degeneration, into one.

\end{enumerate}

In this work, we fully utilize these four data associations and propose the Voxel-SLAM, comprising five key modules: initialization, odometry, local mapping, loop closure, and global mapping. Voxel-SLAM builds upon the efficient LiDAR bundle adjustment of BALM2 \cite{liu2023efficient}, global mapping method of HBA \cite{liu2023large}, the place recognition work of BTC \cite{yuan2024btc}, and the map representation of VoxelMap \cite{yuan2022efficient}. Besides the open-source system itself, the main contributions of Voxel-SLAM are as follows:

\begin{enumerate}
\item Employment of the same map format in all tasks: initialization, odometry, local mapping, loop closure, and global mapping. Voxel-SLAM employs an efficient and versatile adaptive voxel map, which provides sufficient features for different tasks and are well adapted to various scenarios.

\item Robust and fast initialization. Voxel-SLAM only requires short-time data (1s in our realization) to initialize (or re-initialize in the case of system divergence) and can initialize in both stationary and dynamic initial states. The initialization provides accurate states and a consistent map for subsequent modules.

\item Efficient local mapping. Voxel-SLAM uses an efficient LiDAR-inertial BA to refine the local voxel map and states within a sliding window of recent LiDAR scans, enhancing the accuracy and robustness of the system. The local mapping (window size 10) can run at the same rate as its odometry (10 Hz) in real-time on a robot onboard computer with limited computation resources.

\item Loop closure on multiple sessions. Voxel-SLAM has the ability to detect loops in both the current and previous sessions and optimize all the involved scan poses globally.

\item Efficient and accurate global mapping for single or multiple sessions. Voxel-SLAM employs a hierarchical BA to achieve efficient global optimization of scan poses and map consistency. 

\item Full exploitation of four types of data association. Voxel SLAM exploits the short-term and mid-term data associations in its odometry and local mapping, respectively, and the long-term and multi-map data associations in both loop closure and global mapping, to achieve both real-time operation and global map consistency.

\end{enumerate}


\section{Related Works} \label{sec relatedwork}

\subsection{LiDAR(-Inertial) Odometry and SLAM}

As an early LiDAR odometry and mapping framework, LOAM \cite{zhang2014loam} has greatly influenced subsequent works. It includes three modules: feature extraction, scan-to-scan odometry, and scan-to-map mapping. Through the local smoothness of each scanning line, feature points, including planar and edge feature points, are extracted to reduce the computational burden of pose estimation. Then the odometry module uses the feature points to match the current scan to the previous scan to obtain a rough relative pose in real-time. For improved accuracy, the mapping module accumulates historical feature points to build a \textit{k}-d tree and takes the distances of point-to-plane (edge) as the cost to refine the poses from odometry. However, due to the requirement of rebuilding the \textit{k}-d tree, the module of mapping can only run at 1 Hz, compared with the 10 Hz odometry. Following the framework of LOAM, LeGO-LOAM \cite{legoloam2018} takes a ground segmentation method in feature extraction and introduces loop closure to mitigate long-term drift. LOAM-Livox \cite{lin2020loam} adapts the framework into a solid-state LiDAR. The small field of view (FoV) and non-repetitive scanning pattern of the solid-state LiDAR make very few correspondences between two consecutive scans. Therefore, LOAM-Livox only uses the scan-to-map registration for odometry and mapping, which improves the odometry accuracy but increases the computational loading for building the \textit{k}-d tree at each scan.

Incorporating the measurements from an inertial measurement unit (IMU) can provide better initial poses, compensate for motion distortion, and enhance the robustness of estimation for LiDAR SLAM. LIOM \cite{ye2019tightly}, inspired by VINS \cite{qin2018vins}, is one of the first open-source tightly coupled LiDAR-inertial odometry (LIO). However, due to the lack of an efficient BA method, the batch optimization in LIOM is too time-consuming to run in real-time. LiLi-OM \cite{li2021towards}, similar to LIOM using a local map and sliding window optimization for tightly coupled LIO, adapts to solid-state LiDARs and uses an ICP-based loop closure. To ensure real-time performance, LiLi-OM adopts a small optimization window. LIO-SAM \cite{shan2020lio}, discarding the local map with poor efficiency in LIOM, directly feeds the LiDAR odometry results into a factor graph to optimize with IMU preintegration \cite{forster2016manifold}. Thus, LIO-SAM can maintain a larger sliding window and easily fuse with other sensors, such as the GNSS (Global Navigation Satellite System). LINS \cite{qin2020lins} introduces a filter-based method into a LIO system to avoid optimizing multiple states. To reduce the dimension of computing Kalman gain, FAST-LIO \cite{xu2021fast} uses a new Kalman gain. FAST-LIO2 \cite{xu2022fast} as the state-of-the-art LIO system, avoids the feature extraction, and designs a novel incremental \textit{k}-d tree. LTA-OM \cite{zou2024lta} adds the loop closure to FAST-LIO2 with the STD place recognition descriptor \cite{yuan2023std}. Faster-LIO \cite{bai2022faster} replaces the incremental \textit{k}-d tree with parallel sparse incremental voxels in FAST-LIO2, for better efficiency. Point-LIO \cite{he2023point} is a point-by-point LIO framework with a high-frequency odometry output rate and good robustness for aggressive motion and IMU saturation. Recently, there have also been some attempts at combining LIO with LiDAR bundle adjustment (BA) to reduce the LIO drift \cite{li2024ba, tang2024ba} but without any open-source implementation.

Compared with the mainstream open-source LiDAR(-inertial) SLAM systems, Voxel-SLAM has the modules of initialization to enable the system to start at a highly dynamic initial state, local mapping using a LiDAR-inertial BA to enhance the accuracy and robustness, loop closure to detect loop constraints in multiple sessions, and global mapping to further refine the global map. Voxel-SLAM takes full advantage of short-term, mid-term, long-term, and multi-map data associations to achieve real-time estimation and high accuracy mapping.

\subsection{LiDAR Scan Registration Scheme} \label{sec_lidar_registration}

The registration methods for LiDAR point clouds can be divided into two categories: pairwise registration and multiview registration. Pairwise registration only considers estimating a single pose, which is the main registration method in LO and LIO. Due to the sparsity of LiDAR scans, LOAM and its successors employ point-to-plane (edge) registration instead of point-to-point registration, which is used for dense point clouds. They accumulate historical scans into a \textit{points map} to match the latest scan. The drawback of the \textit{points map} is that it takes additional time to form the planes or edges for new scan registration. To avoid this, Suma \cite{chen2019suma++}, LIPS \cite{geneva2018lips}, and VoxelMap \cite{yuan2022efficient} use the \textit{planes map} rather than \textit{points map} to find the corresponding plane for each point in the current scan, achieving better efficiency and accuracy.

Using pairwise registration leads to a cumulative drift, while multiview registration can solve this by registering multiple scans simultaneously. EigenFactor (EF) \cite{ferrer2019eigen} takes the distances of each point to the plane as cost and then transfers the cost into the minimum eigenvalue of a covariance matrix. The cost function is then optimized by a gradient-based method, leading to slow convergence due to the second-order nature of the eigenvalue-type cost. Referencing visual BA, plane adjustment (PA) \cite{zhou2021lidar} considers both poses and plane parameters as optimization variables, employing Schur complement to reduce the dimension brought by plane parameters. The drawbacks are that it takes more iterations to converge when the number of planes increases and the plane parameter has a singularity. For better convergence, BAREG \cite{huang2021bundle} modifies the cost function by adding an extra heuristic penalty term, which could disturb the map consistency terms, especially with real-world noisy point measurements. BALM \cite{liu2021balm} introduces a second-order derivative solver and proves the geometry features can be analytically solved, which takes much fewer steps to converge than the methods mentioned earlier. The main shortcoming of BALM is that it requires enumerating all individual points to compute the Jacobian and Hessian matrix, resulting in high computational complexity due to the dense LiDAR points. To address this problem, BALM2 \cite{liu2023efficient} proposes the \textit{point cluster} to encode the raw points into a compact set of parameters and derive the relevant Hessian and Jacobian matrix. BALM2 has fast convergence speed and high accuracy. Building on BALM2, Liu {\it et al.} proposes a hierarchical bundle adjustment (HBA) framework \cite{liu2023large} for large-scale global mapping.


Voxel-SLAM employs both pairwise registration and multiview registration. In the odometry, Voxel-SLAM leverages an efficient plane map (the VoxelMap \cite{yuan2022efficient}) to achieve efficient scan-to-map registration. In the initialization and local mapping, Voxel-SLAM adopts the BALM2 for the LiDAR(-inertial) BA optimization, considering its efficiency and accuracy. Moreover, to improve the global mapping accuracy, Voxel-SLAM incorporates the efficient HBA framework \cite{liu2023large} after a normal PGO.

\subsection{LiDAR-Based Place Recognition}

Inspired by the normal distributions transform (NDT), Magnusson {\it et al.} \cite{ndtloop} subdivides the point cloud into a regular grid of cells, encodes all cells into a histogram matrix, and detects loop closure by matching these histogram matrices. M2DP \cite{he2016m2dp} projects a 3D point cloud into multiple 2D planes and generates a density signature for each plane, utilizing the left and right singular vectors of these signatures as the descriptors to match. SegMatch \cite{dube2017segmatch} segments the point cloud into semantic features to detect loop retrievals. Giseop Kim {\it et al.}, inspired by shape context \cite{belongie2000shape}, proposes a novel spatial descriptor named Scan Context \cite{kim2018scan}, which encodes a 3D point cloud into a 2.5D information matrix. Their follow-up work, Scan Context++ \cite{kim2021scan}, addresses the issues of lateral invariance and inefficiency of brute-force search in Scan Context by proposing a novel generic descriptor and using sub-descriptors to expedite the loop retrieval. Similar to BoW \cite{galvez2012bags} in visual SLAM, BoW3D \cite{cui2022bow3d} builds the bag of words for 3D features created by Link3D in LiDAR point cloud. In recent years, the deep neural network (DNN) has been introduced to extract local features and encode global descriptors for LiDAR SLAM (LCDNet \cite{cattaneo2022lcdnet}, LoGG3D-Net \cite{vidanapathirana2022logg3d}, LocNet \cite{yin2018locnet}). The main drawback of these learning-based methods is that they are sensitive to the training data and have limited generalization ability. They need to be re-trained when the type of LiDAR or environment is changed. Recently, Yuan {\it et al.} proposed a binary triangle combined (BTC) descriptor \cite{yuan2023std, yuan2024btc}, which introduces a robust binary feature extraction and triangle descriptor matching method with good viewpoint invariance. Besides, BTC can be adapted to any type of LiDAR easily and offers a fast loop retrieval strategy with an accurate 6-DoF relative pose estimation. 

Considering both accuracy and efficiency, Voxel-SLAM incorporates the BTC \cite{yuan2024btc} for loop closure detection. Similar to the voxel map-based scan to map registration and (hybrid) LiDAR bundle adjustment reviewed previously, BTC also uses a voxel map to detect planes for extracting its binary triangle combined descriptors. This makes the adaptive voxel map the unified map structure for all modules of Voxel-SLAM, including the initialization, odometry, local mapping, and global mapping. 

\section{System Overview}

The system overview is shown in Fig. \ref{fig-systemoverview}. The green parts in Fig. \ref{fig-systemoverview} are the different modules in Voxel-SLAM, and the modules in the same thread are grouped in the same gray dashed frame. The blue parts are the data pyramid, and the red parts are the adaptive voxel map used by relevant modules. We define the system state at the end of the $i$-th LiDAR scan as:
\begin{align}
	\mathbf x_i &= \begin{bmatrix}
		\mathbf R_i & \mathbf p_i & \mathbf v_i & \mathbf b^g_i & \mathbf b^a_i
	\end{bmatrix} \label{eq_state}
\end{align}
where $\mathbf R_i \in \mathrm{SO(3)}$, $\mathbf p_i \in \mathbb{R}^3$, and  $\mathbf v_i \in \mathbb{R}^3$ are the rotation, position, and the velocity of the IMU in the world frame, respectively. The world frame is the first IMU frame but having its $z$-axis aligned with the gravity vector, which will be determined in the Initialization module (Section \ref{sec_complete_initialization}). $\mathbf b^g_i \in \mathbb{R}^3$ and $\mathbf b^a_i \in \mathbb{R}^3$ are the bias of gyroscope and accelerator of the IMU in the body frame, respectively.

\subsection{Workflow}

\begin{figure} [t]
	\centering
	\includegraphics[width=1.0\linewidth]{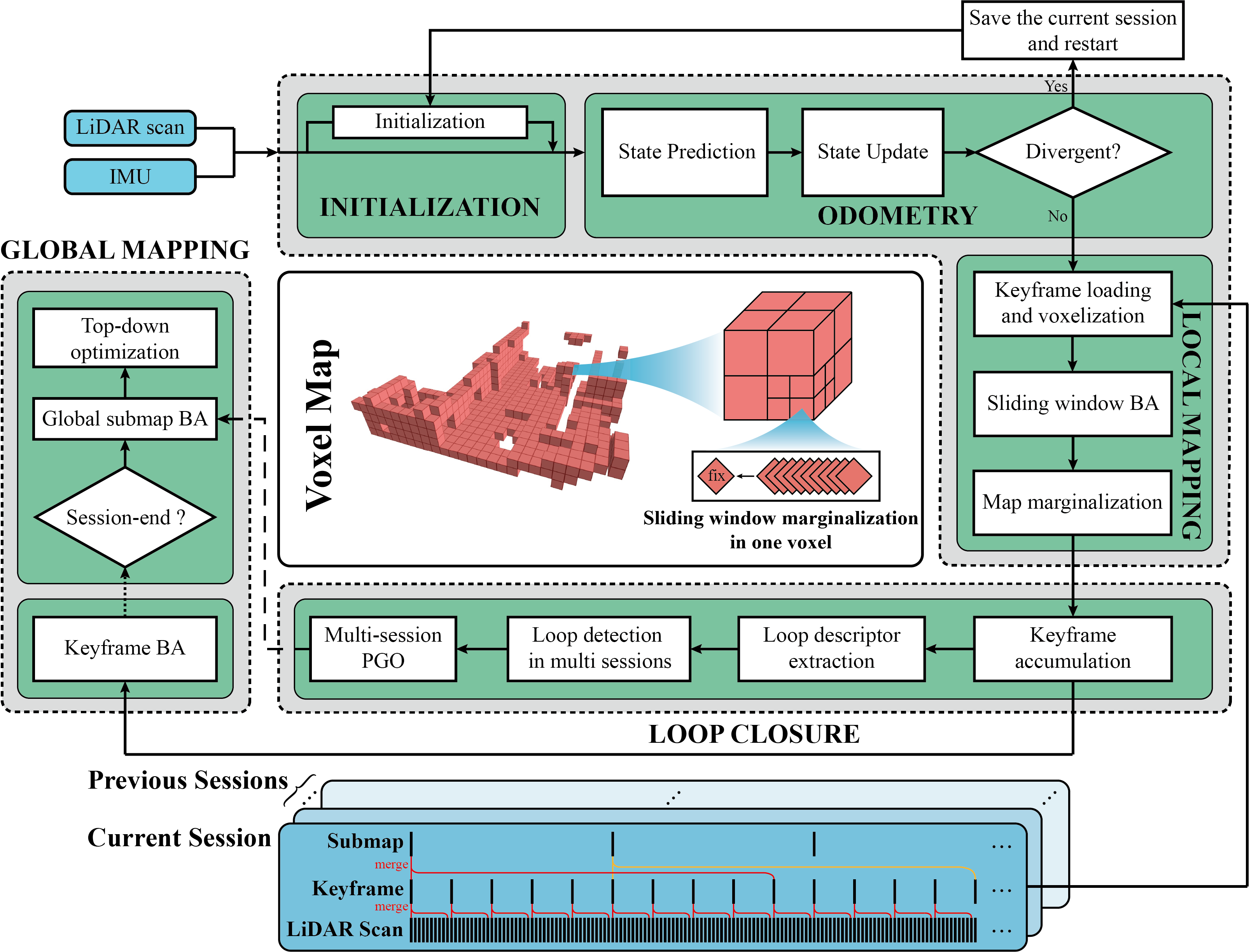}
	\caption{The overview of Voxel-SLAM. The green parts are the different modules of Voxel-SLAM consisting of initialization, odometry, local mapping, loop closure and global mapping. The modules in the same gray dashed box are run in the same thread. The bottom blue parts are the data pyramid of multiple sessions and the center red part is the adaptive voxel map.}
	\label{fig-systemoverview}
\end{figure}

As depicted in Fig. \ref{fig-systemoverview}, Voxel-SLAM consists of five modules running in three parallel threads. Firstly, the system conducts an initialization process (if it has not done so after the start or restart) based on a short time period (1s in our current implementation) of LiDAR scan and IMU data. The initialization employs a specialized LiDAR-inertial BA optimization to estimate the states of all scans, the initial local map, and the gravity vector in the world frame. 
With the initialized state, local map, and gravity vector, the odometry tightly fuses the measurements from LiDAR and IMU to estimate the current state in real-time and detect potential system divergence caused by constant LiDAR degeneration. Subsequently, the local mapping incorporates the current scan into the sliding window and employs a LiDAR-inertial BA to refine all the states within the sliding window and the local map concurrently. The oldest scan within the sliding window will be marginalized to accumulate keyframes, which are then used by the loop closure module to extract loop descriptors and detect loop retrievals across both the current and previous sessions. Upon successful loop detection, the involved poses in the current or previous sessions will be used to construct a pose graph for optimization. Following the pose graph optimization (PGO), the global mapping performs a keyframe BA and merges keyframes into submaps in real time. Upon receiving a session-end signal, the global mapping executes a global BA on all submaps and a top-down optimization to obtain accurate poses of all scans.

\subsection{Data Pyramid} \label{sec_data_pyramid}

A three-level pyramid of data structures is used to contain the input LiDAR point cloud sequence. The bottom level is the raw LiDAR scans, which are directly collected by LiDAR sensors, typically at 10 Hz. The LiDAR scans are used in odometry and local mapping to achieve real-time state estimation. 10 LiDAR scans are merged into one keyframe, which is used in the loop closure to extract the loop descriptor. Finally, 10 keyframes are further merged into a submap to be used in the global mapping module. The merging of LiDAR scans into keyframes and keyframes into submaps is achieved by LiDAR bundle adjustment, which concurrently optimizes the poses of all LiDAR scans (or keyframes) with respect to the first scan (or keyframe) in the respective merging window. Moreover, in the process of merging keyframes into submaps, two consecutive merging windows share five overlapped keyframes (i.e., a sliding window with a size of 10 and a stride of 5) to increase the portion of co-visible areas among the two consecutive submaps. On the other hand, the merging window from scans into keyframes has no overlap (a sliding window with a size of 10 and no stride), considering that the long LiDAR measuring distances have already ensured large overlaps among scans in two consecutive keyframes.  


\subsection{Adaptive Voxel Map} \label{sec_adaptive_voxel_map}

The adaptive voxel map plays a crucial role in all modules to extract plane features and provide feature associations across multiple scans, keyframes, or submaps. The voxel map maintains the plane features of different sizes in the environments. To do so, it splits the space into uniform voxels (i.e., root voxels), each with size $L_r$. The uniform-sized voxels are organized by a Hash table, where each Hash key maps each LiDAR point to the corresponding voxel by its point location. A voxel contains an octree data structure of multiple layers, where each leaf node represents a plane. Leaf nodes at different layers have different sizes, representing plane features of varying sizes. 

An adaptive voxel map is constructed as follows. For each scan, keyframe, or submap, its points are distributed into the corresponding voxels by their positions. If the points in a voxel are on the same plane (the ratio between the minimum and the second minimum eigenvalue of the point covariance matrix is less than a specified value), the voxel is considered a plane voxel and saved with the LiDAR points for subsequent use; otherwise, the voxel is recursively split into smaller sub-voxels until the points are on the same plane or the termination condition is met (e.g., reaching the minimum sub-voxel size or number of points).

The adaptive voxel map structure is used in different modules for different purposes. First, a local adaptive voxel map within the distance $L_m$ around the LiDAR is shared by the modules of initialization, odometry, and local mapping to estimate the states of LiDAR scans in real time (see the voxel map in Fig. \ref{fig-systemoverview}). The initialization first obtains a number of initial scans to initialize the voxel map. The odometry aligns the current scan with the planes in the voxel map to estimate the current state. The local mapping module slides the local map and refines the state of the current and recent scans through an efficient LiDAR-inertial bundle adjustment (BA) optimization. Besides the local voxel map, two other adaptive voxel map structures are also used: one used by the loop closure module, the BTC \cite{yuan2024btc}, for descriptor extraction in a keyframe, and the other by the global mapping module, the HBA \cite{liu2023large}, for plane feature extraction and association across different keyframes and submaps.


\section{LiDAR-Inertial BA Optimization} \label{sec_li_ba_opt}

The LiDAR-inertial BA optimization is used in the initialization and local mapping modules to estimate multiple scan states simultaneously. Compared with the filter-based estimation of LIO methods \cite{xu2021fast, xu2022fast}, the estimation of LiDAR-inertial BA is more accurate and robust due to the use of longer (mid-term) data association. However, a disadvantage of the BA method is its increased computation time, preventing its use in most existing LiDAR(-inertial) odometry or SLAM systems. To solve this problem, BALM2 \cite{liu2023efficient} introduces an efficient LiDAR BA method by: (1) solving the geometric features (i.e., plane and line) analytically, which eliminates the feature parameters from the BA optimization, hence drastically reducing the optimization dimension; (2) introducing the \textit{point cluster} data structure, which aggregates all points into a compact representation, hence avoiding the enumeration of each raw point in the BA; (3) deriving the analytical second-order derivatives of the BA optimization, which enable the development of an efficient second-order solver. The three techniques make the LiDAR BA optimization highly efficient for real-time implementation.

\begin{figure} [t]
	\centering
	\includegraphics[width=1.0\linewidth]{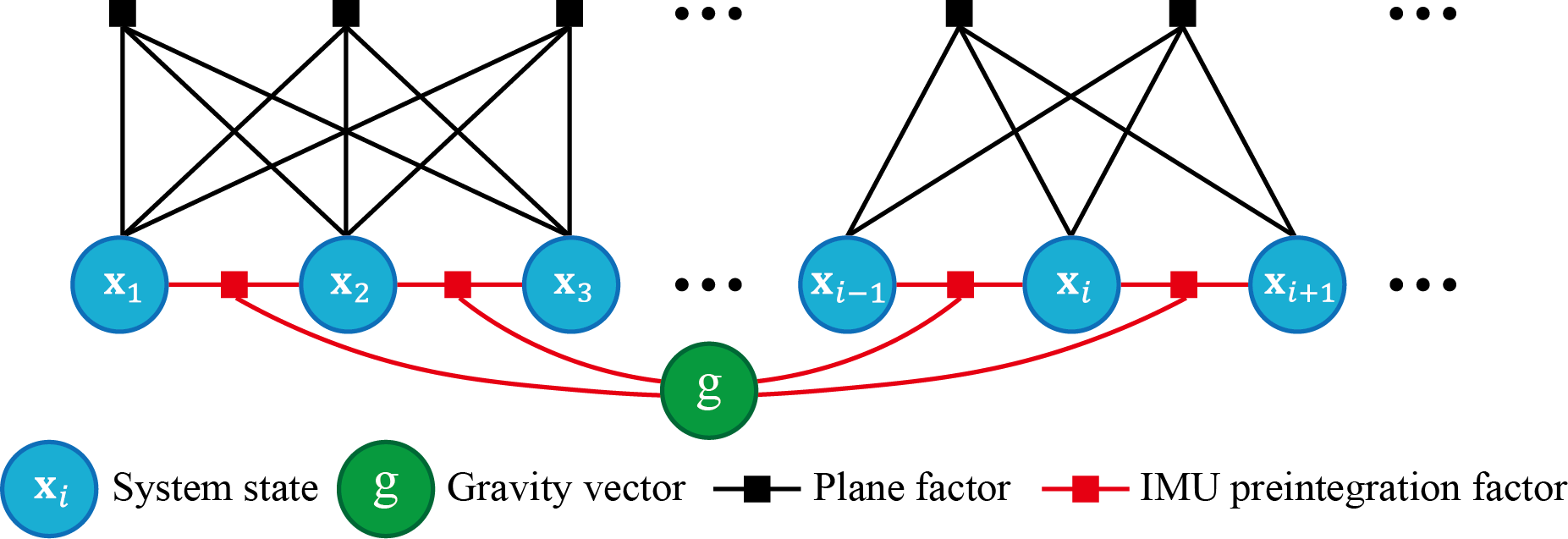}
	\caption{The factor graph representation of the proposed LiDAR-inertial bundle adjustment.}
	\label{fig-factorgraph}
\end{figure}

We employ the method of BALM2 and further combine it with IMU preintegration \cite{forster2016manifold} to formulate a LiDAR-inertial BA optimization to estimate multiple states concurrently as well as the gravity vector if needed (see the factor graph representation in Fig. \ref{fig-factorgraph}). The cost function is:
\begin{align}
\arg\min_{\mathcal X} \bigg(
\frac{1}{2} & \sum_{i=1}^{N-1} \left\| 
\mathbf r_{i,i+1}(\mathcal X)
\right\|^2_{\boldsymbol\Sigma_{i,i+1}^{-1}}
+ \sum_{j=1}^{M} \lambda_{j}^{\text{min}} (\mathcal X)
\bigg) \label{eq-liba-cost}
\\
\mathcal X & = 
\begin{bmatrix}
	\mathbf x_1,\ \mathbf x_2,\ \cdots,\ \mathbf x_{N},\ \mathbf g
\end{bmatrix}
\end{align}
where $\mathcal X$ is the state vector consisting of individual state $\mathbf x_i$ and the gravity vector $\mathbf g$, $\mathbf r_{i,i+1}(\mathcal X)$ is the residual of IMU preintegration between $i$-th and ($i$+1)-th states with its covariance matrix $\boldsymbol\Sigma_{i,i+1}$, $\lambda_{j}^{\text{min}}$ is the minimum eigenvalue of the points covariance matrix associated to the $j$-th plane feature, which accounts for the contribution of LiDAR BA factor (see details in \cite{liu2023efficient}), and $N$ and $M$ are the number of involved states and plane features, respectively. The residual of IMU preintegration $\mathbf r_{i,j} (i<j)$ directly follows \cite{forster2016manifold}:
\begin{align}
&\mathbf r_{i,j} = 
\begin{bmatrix}
    \mathbf r_{\Delta \mathbf R_{ij}},\ \mathbf r_{\Delta \mathbf p_{ij}},\ \mathbf r_{\Delta \mathbf v_{ij}},\ \mathbf r_{\Delta \mathbf b^g_{ij}},\ \mathbf r_{\Delta \mathbf b^a_{ij}}
\end{bmatrix} \notag
\\
& \mathbf r_{\Delta \mathbf R_{ij}} = 
\text{Log}(\Delta \tilde{\mathbf R}_{ij}^T(\mathbf b_i^g) \mathbf R_{i}^T \mathbf R_{j}) \notag
\\
& \mathbf r_{\Delta \mathbf p_{ij}} = \mathbf R_{i}^T (\mathbf p_{j} - \mathbf p_{i} - \mathbf v_i \Delta t_{ij} - \frac{1}{2} \mathbf g \Delta t_{ij}^2) - \Delta \tilde{\mathbf p}_{ij} (\mathbf b_i^g, \mathbf b_i^a) \notag
\\
& \mathbf r_{\Delta \mathbf v_{ij}} = \mathbf R_{i}^T (\mathbf v_{j} - \mathbf v_{i} - \mathbf g \Delta t_{ij}) - \Delta \tilde{\mathbf v}_{ij} (\mathbf b_i^g, \mathbf b_i^a) \notag
\\
& \mathbf r_{\Delta \mathbf b^g_{ij}} = \mathbf b^g_j - \mathbf b^g_i \qquad \mathbf r_{\Delta \mathbf b^a_{ij}} = \mathbf b^a_j - \mathbf b^a_i \label{eq_residual_imu_pre}
\end{align}
where $\mathbf R_i$, $\mathbf p_i$, $\mathbf v_i$, $\mathbf b_i^g$, and $\mathbf b_i^a$ constitute the state $\mathbf x_i$ as defined in (\ref{eq_state}), and the Jacobian matrices of $\mathbf r_{i,j}$ about state $\mathbf x_i$ are the same with \cite{forster2016manifold}. $\Delta\tilde{\mathbf R}_{ij}^T(\mathbf b_i^g)$, $\Delta\tilde{\mathbf p}_{ij} (\mathbf b_i^g, \mathbf b_i^a)$, and $\Delta\tilde{\mathbf v}_{ij} (\mathbf b_i^g, \mathbf b_i^a)$ are the corrected preintegrated rotation, position, and velocity using first-order expansion in \cite{forster2016manifold}. $\Delta t_{ij}$ is the interval between time $i$ and $j$. $\text{Log}: \text{SO}(3) \rightarrow \mathbb R^3$ maps from the Lie group to the vector space. $\mathbf g \in \mathbb R^3$ is the gravity vector. 

It is noted that the gravity vector is treated as a vector in $\mathbb R^3$ with scale subject to optimization instead of $\mathbb S^2$ with fixed scale as in visual SLAM. This is because the measurement of LiDAR has accurate scale information, making it possible to determine the scale from the data. In contrast, the visual measurements is well-known of lacking the scale information, making a fixed-scale gravity necessary in most visual SLAM systems \cite{qin2018vins, campos2021orb}. 

The cost function in (\ref{eq-liba-cost}) is optimized iteratively using a second-order solver, the LM solver. To speed up the optimization process, we derived the Jacobian and Hessian of the cost function in analytical form. The details of Jacobian, Hessian, and the solver can be found in Appendix \ref{iterative BA}.


\section{Initialization} \label{sec_initialization}

Many filter-based LIO systems \cite{xu2021fast, xu2022fast, bai2022faster, he2023point} assume the sensor is initially static. For non-static initial states (e.g., restart in the middle of a sequence), the static assumption will propagate incorrect states, leading to errors in distortion correction of LiDAR scans and feature association in scan registration. This can significantly impact the accuracy of the earlier time of the state estimation and result in an inconsistent (or even incorrect) initial map that further degrades the subsequent odometry. To address this limitation, we propose a robust and efficient initialization stage, based on the previous LiDAR-inertial BA optimization, to provide robust estimation of the initial states, the gravity vector, and a consistent initial local map even in the presence of a non-static initial state. Besides, the module can confirm if this initialization is successful and re-align the gravity vector to the $z$-axis of the world frame.


The initialization is conducted on the first $N$ LiDAR scans along with corresponding IMU measurements. To be specific, it first runs the LIO method in \cite{xu2021fast} in real time to estimate roughly the states of the first $N$ scans and the gravity vector. The estimated states and the gravity vector serve as the initial values for the subsequent coarse-to-fine voxelization and BA optimization (Section \ref{coarse-fine-voxelization}). The optimized states and gravity vector are used to determine whether the initialization is successful (Section \ref{sec_criteria_initialization}). If yes, the initialization is considered complete (Section \ref{sec_complete_initialization}); otherwise, the system will collect the next $N$ LiDAR scans and repeat the above initialization process until success.


\subsection{Coarse-to-Fine Voxelization and BA Optimization} \label{coarse-fine-voxelization}

Given a rough initial state (e.g., estimated by LIO), the coarse-to-fine voxelization and BA optimization aim to refine the state and corresponding map based on the LiDAR-initial BA optimization outlined in Section \ref{sec_li_ba_opt}. Inaccurate initial states may introduce large discrepancies in the registered point cloud map, leading to false positive or false negative plane associations in the voxelization process (see Section \ref{sec_adaptive_voxel_map}). The wrong plane associations then cause the errors of BA optimization in turn. To overcome this issue, we propose a coarse-to-fine voxelization and BA optimization where the criterion for determining a plane feature in the adaptive voxelization process (Section \ref{sec_adaptive_voxel_map}) is tightened gradually. At first, the criterion is looser, recalling plane features even under large initial state errors. Based on the recalled plane features, the LiDAR-inertial BA optimization is performed. The refined state is then used to compensate for the motion distortion in each scan and to construct the next round of the adaptive voxel map, which has a tighter criterion for plane association. The voxelization and BA optimizations are conducted many rounds until convergence, where in each round the plane criterion is tightened from the previous round. The coarse-to-fine process is considered to converge when the decrement in the cost value, from the last round to the current one, is less than a certain value.

\subsection{Criteria for Initialization Success} \label{sec_criteria_initialization}

Three criteria are used to determine if the initialization is successful: (1) the coarse-to-fine process should converge within a certain number of rounds; (2) the magnitude of the optimized gravity vector should be close to the value of 9.8 $\text{m/s}^2$; (3) the initial voxel map, as constructed from the optimized state, should contain plane constraints in three linearly independent directions. This degeneration assessment is detailed in Appendix \ref{degeneration judge}. If the degeneration occurs in the final BA optimization, it means the involved states in the initialization are not constrained well with sufficient plane features and could have large estimation errors. 

\subsection{Completion of Initialization} \label{sec_complete_initialization}

Upon successful initialization, we obtain the accurate gravity vector and states with undistorted point clouds from the first $N$ scans. The gravity vector is then used to align to the $z$-axis of the world reference frame by a rotation matrix $\mathbf R_{\mathbf g}^{z}$, and the states are also rotated by the matrix. Afterwards, the gravity vector will be fixed in the subsequent modules. The adaptive voxel map built from the optimized states is serving as the initial local map for the subsequent odometry module. The initialization is conducted only once for each start (or restart), meaning that the subsequent LiDAR scans and IMU measurements will be fed to the odometry directly.

\section{Odometry}


In Voxel-SLAM, the odometry exploits the short data association through scan-to-map registration. The aim of the odometry is to estimate the current state with minimal time delay, which is necessary for other tasks such as planning and control. The estimated state also provides a relatively accurate initial value for the subsequent local mapping module. 

\subsection{State Prediction and Update}

To conduct state estimation on the local adaptive voxel map, we use the method in \cite{yuan2022efficient} for scan-to-map correspondence and the extended Kalman filter (EKF) in \cite{xu2021fast} to tightly couple LiDAR and inertial measurements. Specifically, upon synchronizing the LiDAR scan and IMU measurement, the odometry predicts the state through IMU propagation and, meanwhile, compensates for the motion distortion of the LiDAR scan. The points in the undistorted scan are down-sampled spatially at a prescribed resolution $L_d$ and transformed from the LiDAR body frame to the IMU body frame by the pre-known LiDAR-IMU extrinsic parameter. Each point in the scan computes a covariance matrix by the method in \cite{yuan2022efficient} and is projected into the world frame to match the plane in the voxel map with the maximum probability. The point-to-plane distances are utilized as the system measurements, which update the IMU-predicted state within an error-state iterative Kalman filter \cite{xu2021fast}. The estimated state and LiDAR scan will be published to other modules for further processing.

\subsection{Detection of System Divergence} \label{sec_system_divergence}

When a robot encounters a degenerate environment that lacks sufficient geometric features, the state estimation becomes ill-posed. This would cause the SLAM system to diverge and fail, which should be actively monitored and handled. In Voxel-SLAM, we detect the potential divergence of the system after each scan registration. After the state estimation in the odometry, the planes used in the scan-to-map correspondence are collected to assess the degeneration by the method in Appendix \ref{degeneration judge}. An occasional degeneration in a scan is not considered a divergence due to the relatively accurate IMU prediction over a short period. Constant degeneration over consecutive scans will be considered a divergence. In such cases, the current data session is saved, and the whole system will restart from a new origin that is far apart from the current session to prevent the points in the next session from overlapping with those in the current session. The restarting will also trigger a new initialization to obtain an initial state and local map for the next session.

\section{Local Mapping} \label{sec-local-mapping}

The local mapping, leveraging the mid-term data association, refines the local voxel map and all recent states within a sliding window. Compared with the odometry, the local mapping is more accurate and robust since its LiDAR-inertial BA takes advantage of a longer period of measurements from LiDAR and IMU for state estimation. 

\subsection{Keyframe Loading and Voxelization} \label{sec_voxelization}

When the local mapping receives the IMU measurements and the current LiDAR scan with its estimated state from the odometry, the points in the scan are distributed into corresponding voxels by their positions in the world frame. Meanwhile, the IMU measurements are preintegrated according to \cite{forster2016manifold} to obtain the IMU preintegration factors. Due to only maintaining a local map within the distance $L_m$ around the current LiDAR position, the voxel map cannot ensure long-term data association in the case of revisiting a place existing in the current or previous sessions. To establish long-term association, we dynamically load the keyframes {that are temporally distant (e.g., $\geq 1$ min to avoid recent keyframes) but spatially close (e.g., $\leq 10$ m) to the current scan}, a technique proved to be effective in \cite{zou2024lta}. The points in these keyframes are distributed into corresponding voxels, serving as fixed historical map constraints for the sliding window BA optimization.

When distributing points from the current scan or dynamically-loaded keyframes into voxels, we keep track of the voxels that have newly added points. These voxels are enumerated, where all the contained points, either from the current scan, keyframes, or existing points, go through an adaptive voxelization (Section \ref{sec_adaptive_voxel_map}) until each leaf node has a valid plane. In each leaf node, all points are encoded into a compact structure, \textit{point cluster} \cite{liu2023efficient}, and different scans within the sliding window will have different point clusters. The remaining points in the leaf node are termed as ``fix" points and serve as the map constraints for the sliding window optimization (these points may be from the dynamically-loaded keyframes or previously-marginalized scans, as in Section \ref{sec_map_marginalization}). These ``fix" points are encoded into a single point cluster, which, along with the point cluster of each scan in the sliding window, will be used directly for BA optimization without enumerating each individual point.

\subsection{Sliding Window BA Optimization}

\begin{figure} [t]
	\centering
	\includegraphics[width=1.0\linewidth]{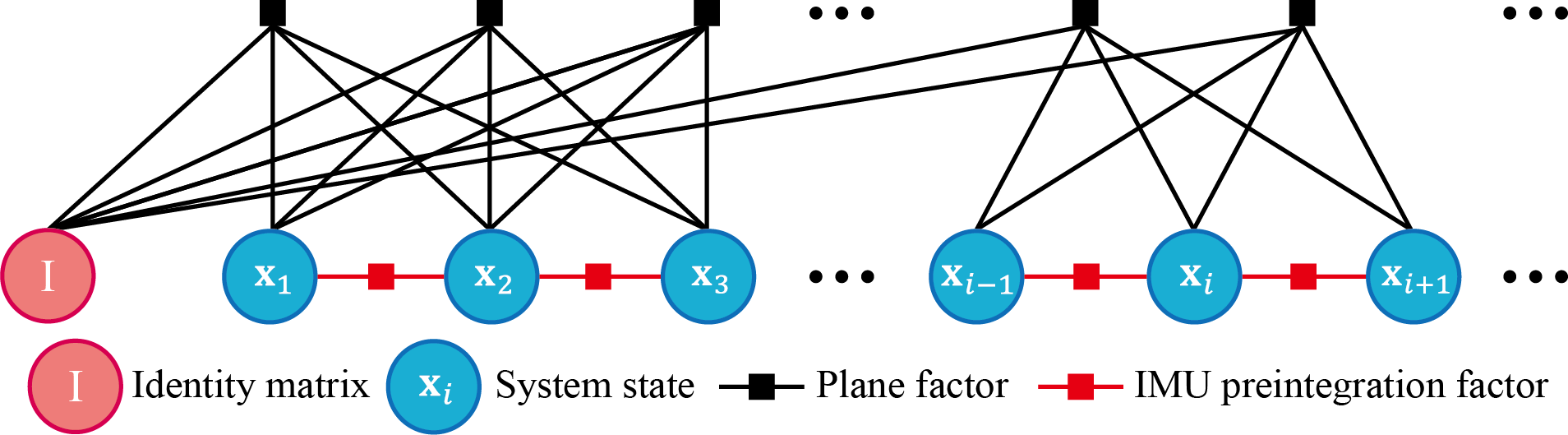}
	\caption{The factor graph representation of the LiDAR-inertial bundle adjustment used in local mapping. The ``identity matrix" represents the map constraints from the ``fix" points (represented in the form of point cluster), which are out of the sliding window optimization and are represented in the world frame.}
	\label{fig_factorgraph_nogravity}
\end{figure}

After obtaining the point cluster associations and IMU preintegrations, the sliding window BA optimization employs the LiDAR-inertial BA optimization in Section \ref{sec_li_ba_opt} to refine all the states within the sliding window. One difference between the sliding window BA optimization and that in Section \ref{sec_li_ba_opt} is that the sliding window BA optimization incorporates the constraints from the ``fix" points in each voxel, serving as map constraints to ensure the accuracy of the estimation in the world frame. Another difference is that the gravity vector is fixed, considering that the initialization has estimated the gravity vector well. Fig. \ref{fig_factorgraph_nogravity} is a factor graph representation of the sliding window LiDAR-inertial BA optimization.

\subsection{Map Marginalization} \label{sec_map_marginalization}


After the BA optimization, the point clusters and plane parameters (plane centers and normal vectors) in the local voxel map are updated by the refined states. The updated planes will be used in the next scan-to-map correspondence of the odometry module. Moreover, the oldest scan within the sliding window will be marginalized. That is, the oldest scan is removed from the sliding window, and its points in each leaf node are merged into the ``fix" point cluster as map constraints. After the marginalization, the oldest scan and its pose will be sent for loop closure for further processing. Moreover, the voxels whose leaf nodes have no points corresponding to the states within the sliding window or whose distances from the current position are larger than the local map size $L_m$, will be removed from the local map to reduce memory use.



\section{Loop Closure} \label{loop closure}


Loop closure serves two important purposes within our framework. Firstly, by exploiting long-term data association, it effectively mitigates drift by detecting previously visited locations and corrects the cumulative errors by pose graph optimization (PGO). Secondly, loop closure matches the current session with previous sessions in the same world frame to achieve multi-map data associations. This process enables the integration and alignment of data collected from different mapping sessions, facilitating the creation of a unified map of the environment.

\subsection{Keyframe Generation and Loop Detection} \label{sec_loop_detection}

A scan, after being marginalized from the sliding window in the local mapping, will be pushed into the pose graph by the loop closure module. $N$ consecutive marginalized scans accumulate into a frame, which is then selected as a keyframe if the moving distance and rotation angle between the current frame and latest keyframe are larger than predetermined thresholds (e.g., $0.5$ m and $5$ deg). The keyframe is used to extract a BTC descriptor and to match candidate loop keyframes in the current and previous sessions, as detailed in \cite{yuan2024btc}. The keyframe is meanwhile sent to the thread for global mapping for further processing.

\begin{figure} [t]
	\centering
	\includegraphics[width=1.0\linewidth]{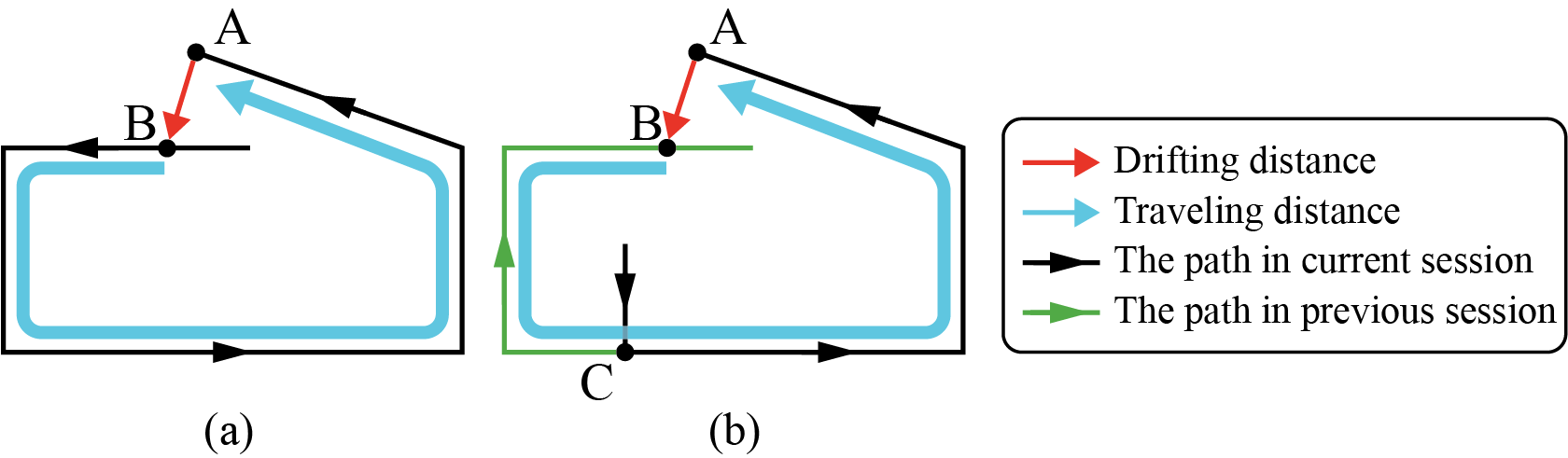}
	\caption{Drifting and traveling distances. (a) The candidate loop keyframe is in the current session. The traveling distance is the distance accumulated along the current session from the candidate loop keyframe B to the current keyframe A. (b) The candidate loop keyframe is in a previous session. The traveling distance is the distance accumulated along the previous session from the candidate loop keyframe B to the previous loop keyframe C and then along the current session from C to the current keyframe A. For both cases, the drifting distance is the relative distance between the current keyframe A and the candidate loop frame B, which is computed from the loop detection method BTC \cite{yuan2024btc}.}
	\label{fig-loopdrift}
\end{figure}


Upon identifying a loop closure candidate, it is essential to verify its authenticity to avoid false-positive detection. In addition to the built-in geometric verification of BTC \cite{yuan2024btc}, we take two extra criteria to enhance the reliability of loop detection. (1) The point-to-plane alignment of the current keyframe with the detected candidate should contain plane constraints in three linearly independent directions with the criteria in Appendix \ref{degeneration judge}. (2) The ratio of drifting distance to traveling distance should be small enough (e.g., $\leq 1\%$). This criterion applies to the cases where the candidate loop keyframe is within the current session (see Fig. \ref{fig-loopdrift}(a)) or within the previous session for the second (or more) time (see Fig. \ref{fig-loopdrift}(b)). This criterion is based on the observation that, in a well-performing SLAM system, the drift after a fixed distance of travel is basically within a certain value. If the ratio is too large, this loop retrieval is most likely a mismatch. Moreover, if the candidate loop keyframe is within the previous session for the first time, the second criterion would not apply, and the first criterion as well as the geometric verification within BTC \cite{yuan2024btc} should be more stringent considering the detrimental effect on the subsequent operation of the system if the first loop detection across different sessions is wrong.

\subsection{Pose Graph Optimization and Map Rebuilding} \label{sec_pgo_map_rebuild}


\begin{figure} [t]
	\centering
	\includegraphics[width=1.0\linewidth]{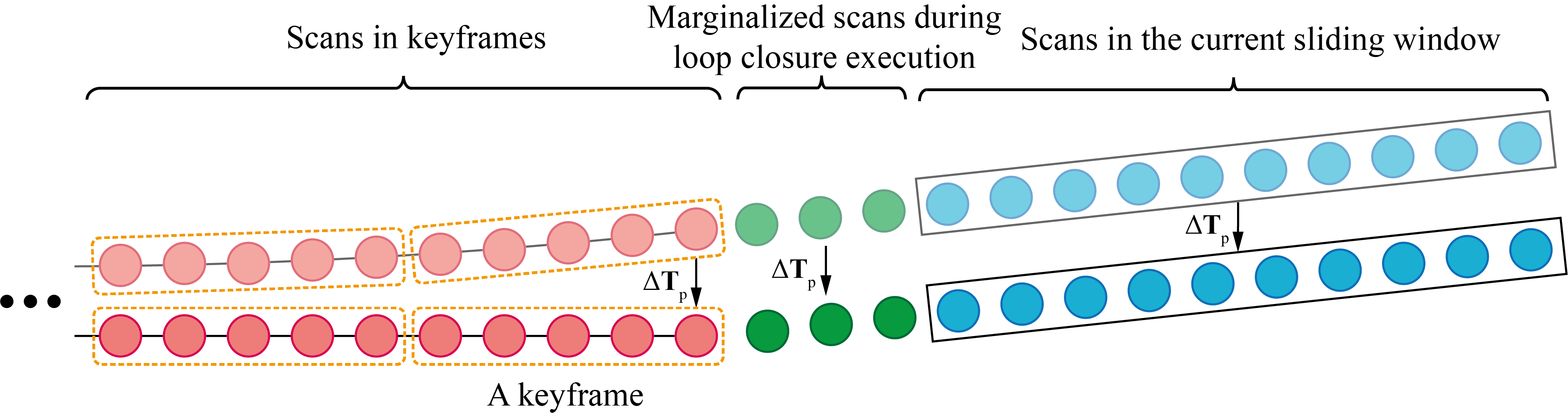}
	\caption{The scans before and after a PGO. The red nodes are the scans that have been added to the pose graph. The green nodes are the marginalized scans from the local mapping during the loop closure execution, hence having not been added to the pose graph yet. The blue nodes are the scans in the current sliding window of the local mapping. The nodes at the top and bottom are the scans before and after PGO, respectively. $\Delta \mathbf T_\text{p}$ is the pose correction of the last scan in the pose graph and used to correct the subsequent scan beyond the pose graph.}
	\label{fig_correction_propagation}
\end{figure}

Upon a successful loop detection, if the detected loop keyframe is in the previous session for the first time, the current session is connected to the previous session by aligning all poses in the current session into the world frame of the previous session and concatenating the pose graph of the two sessions. Otherwise, the detected loop keyframe is in the previous session for the second (or more) times or in the current session, either case will cause a loop within the pose graph, which subsequently triggers a pose graph optimization (PGO) via GTSAM \cite{dellaert2012factor}. 


After updating scan poses via PGO, the poses of keyframes and submaps are also synchronized by their corresponding updated scan poses, as shown in the data pyramid of Fig. \ref{fig-systemoverview} (see the red nodes in Fig. \ref{fig_correction_propagation}). Moreover, the poses of the scans beyond the pose graph should also be updated to ensure the consistency of the system. These scans consist of (1) marginalized scans during the loop closure execution (see the green nodes in Fig. \ref{fig_correction_propagation}); and (2) all scans in the current sliding window, which has slid forward after the last scan was added to the pose graph (see the blue nodes in Fig. \ref{fig_correction_propagation}). These two types of scans have their poses all corrected by the transformation, $\Delta \mathbf T_\text{p}$, which is the pose correction of the last scan in the pose graph as shown in Fig. \ref{fig_correction_propagation}. The two types of scans with updated poses, along with all scans in the latest five keyframes in the pose graph, are then used to build a new adaptive voxel map. In each leaf node of the new voxel map, points belong to the keyframes, and marginalized scans are the ``fix" points serving as the map constraints, so they are encoded into one single point cluster. In contrast, points belonging to scans in the current sliding window are encoded into point clusters separately (see Section \ref{sec_voxelization}). Finally, the new voxel map will replace the original voxel map to be used by the odometry and local mapping in the future. To save computational resources, the PGO and map rebuilding are conducted only when the drifting distance in Section \ref{sec_loop_detection} is larger than a predetermined threshold (e.g., 0.1m).

\section{Global Mapping} \label{sec_global_mapping}

The point cloud map from the loop closure, particularly in multi-session SLAM, may exhibit inconsistencies, due to the sole consideration of pose constraints in pose graph optimization (PGO), which does not capture direct map consistency constraints. To address this, we introduce the global mapping to further refine the poses and maps from the loop closure, employing a hierarchical global BA method \cite{liu2023large} to mitigate excessive optimization dimensions.

Based on the data pyramid in Fig. \ref{fig-systemoverview}, the global mapping thread encompasses three procedures. Firstly, after receiving keyframes from the loop closure thread, the global mapping thread executes a bundle adjustment (BA) on a window of keyframes in real-time. The window has a size of 10 and slides with a stride of 5. In each sliding window BA, the pose of the first keyframe is fixed at the PGO result, and the rest nine keyframe poses are optimized concurrently in the optimization. After that, the 10 keyframes in each window are merged into a submap. Secondly, once the current session ends (e.g., data collection ends or the system restarts), a global BA on submaps is conducted. Considering that the scan pose returned by PGO may not render sufficient global map consistencies, the coarse-to-fine voxelization and BA optimization method in Section \ref{coarse-fine-voxelization} is adopted for the global submap BA, where the IMU pre-integration factors and gravity optimization are removed. The BA optimization from keyframes to submap, and then from submap to global map, is known as the bottom-up BA optimization in \cite{liu2023large}, which significantly reduces the optimization dimensions compared to the global BA directly on keyframes or scans. Finally, a top-down optimization \cite{liu2023large} with PGO is employed to improve the global consistency.

\section{Experimental Results} \label{sec experiments}

We perform five sets of experiments to validate the accuracy and versatility of our system, as follows:
\begin{enumerate}
\item Evaluating the robustness, accuracy, and time consumption of the initialization.
\item Single-session SLAM benchmark comparison with other odometry and SLAM methods.
\item Multi-session SLAM experiments.
\item Multi-session SLAM with Online relocalization on a computation-limited computer.
\item Computational time of experiments 2)-4) above.
\end{enumerate}

We use three public datasets with significant differences: Hilti \cite{zhang2022hilti, nair2024hilti}, MARS-LVIG \cite{li2024mars}, and UrbanNav \cite{hsu2023hong}, which include a total of 30 sequences, and a private dataset with 2 sequences. {The details of all the 32 sequences about their duration and distances are listed in Table VIII of the Appendix \ref{appen_sequence_list}.} The Hilti is a SLAM dataset collected in indoor and outdoor structured construction environments. We use their handheld sequences, which are collected by a Hesai XT-32 LiDAR and Bosch BMI085 IMU at 400 Hz, and can export the absolute trajectory error (ATE) results from their website. The MARS-LVIG collects the data on an unmanned aerial vehicle (UAV) from a downward-looking Livox Avia LiDAR (with a built-in IMU BMI088 at 200 Hz) at a height of about 100 meters. The UrbanNav is an urban robot-car dataset with the Xsens-MTI-30 IMU at 400 Hz and a HDL-32E LiDAR. Our private dataset is collected on a light-weight handheld device, as shown in Fig. \ref{fig-relc_equipment}, assembled with a Livox Avia LiDAR and its internal IMU. In all the above sequences, the LiDAR runs at 10 Hz.

The first three experiments mentioned above were run on a laptop computer with an Intel i7-10750H CPU at 3.5 GHz and 32 GB of memory. The fourth one runs on an onboard computer with an Intel i3-N305 CPU at 3.0 GHz and 16 GB of memory. The root voxel size and spatial down-sampling resolution for the scan are $L_r = 1$ m and $L_d = 0.1$ m in all indoor environments, $L_r = 2$ m and $L_d = 0.25$ in all outdoor scenario, and $L_r = 4$ m and $L_d = 0.5$ m in the high-altitude downward-looking environment, regardless of the datasets or sequences. The local voxel map size is $L_m=1,000$ m and the maximum layer of the map is $l_{max} = 3$. The minimum number of points on a plane is $N_{\text{min}} = 5$, and the plane criteria is {$\frac{\lambda_3}{\lambda_2} < \frac{1}{16}$}, where $\lambda_l\ (l=1,2,3)$ is the eigenvalue of the plane covariance matrix with $\lambda_1 \geq \lambda_2 \geq \lambda_3$. The sliding window size is $N = 10$ for the initialization, LiDAR-inertial BA, and keyframe BA. These parameters are kept the same across all datasets and sequences. More details of the experiments can be found in the video\footnote{\url{https://youtu.be/Cg9W01aIUzE}}.

\subsection{Initialization}

\begin{figure} [t]
	\centering
	\includegraphics[width=1.0\linewidth]{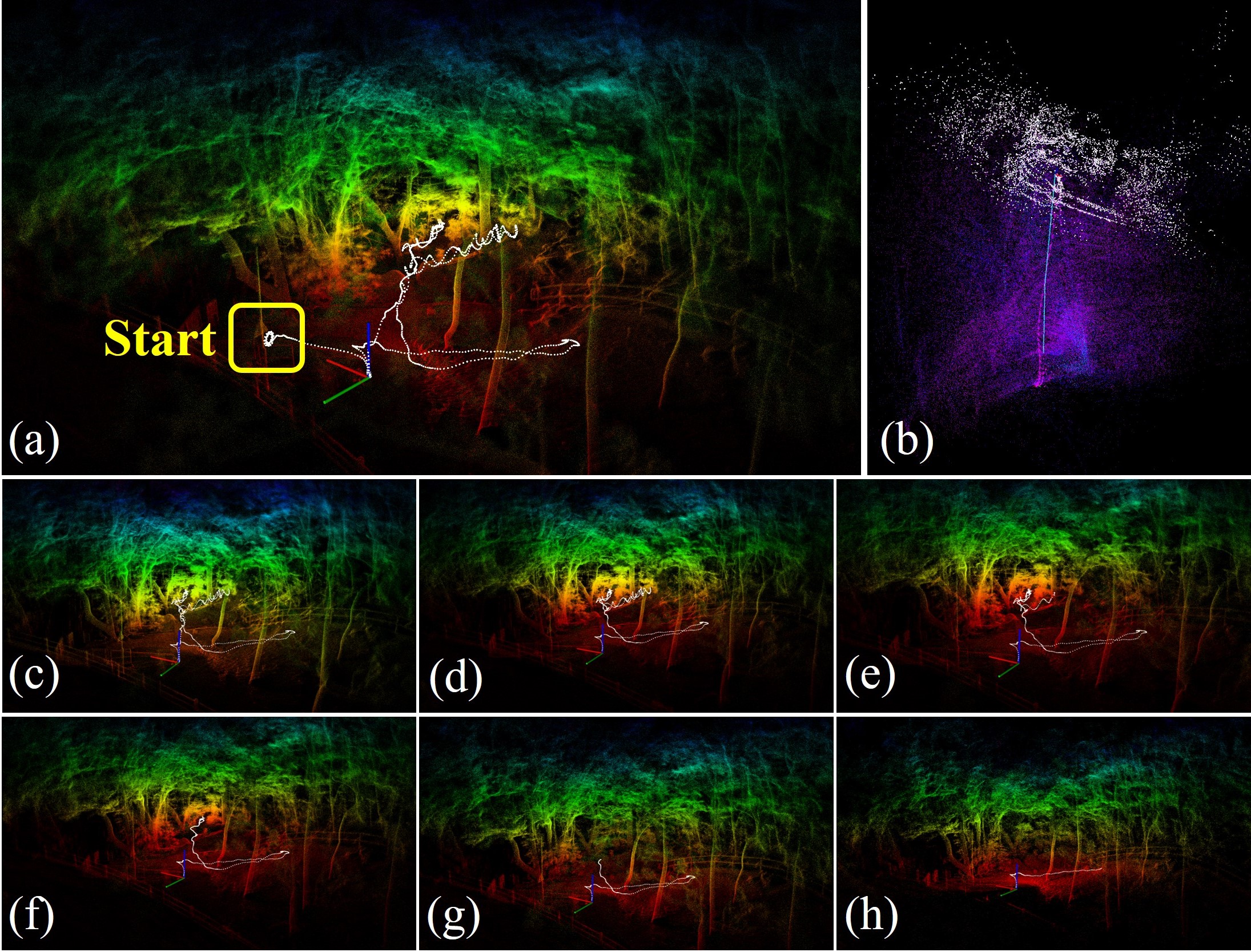}
	\caption{A sequence with large initial motion. (a) Voxel-SLAM initializes successfully and completes the whole sequence. (b) FAST-LIO2 is divergent due to the violent initial movement. (c)-(h) Voxel-SLAM successfully initializes when starting from the $1/7$ to $6/7$ of the sequence, which testifies the robustness of initialization.}
	\label{fig-violent-init}
\end{figure}

\begin{figure} [t]
	\centering
	\includegraphics[width=0.95\linewidth]{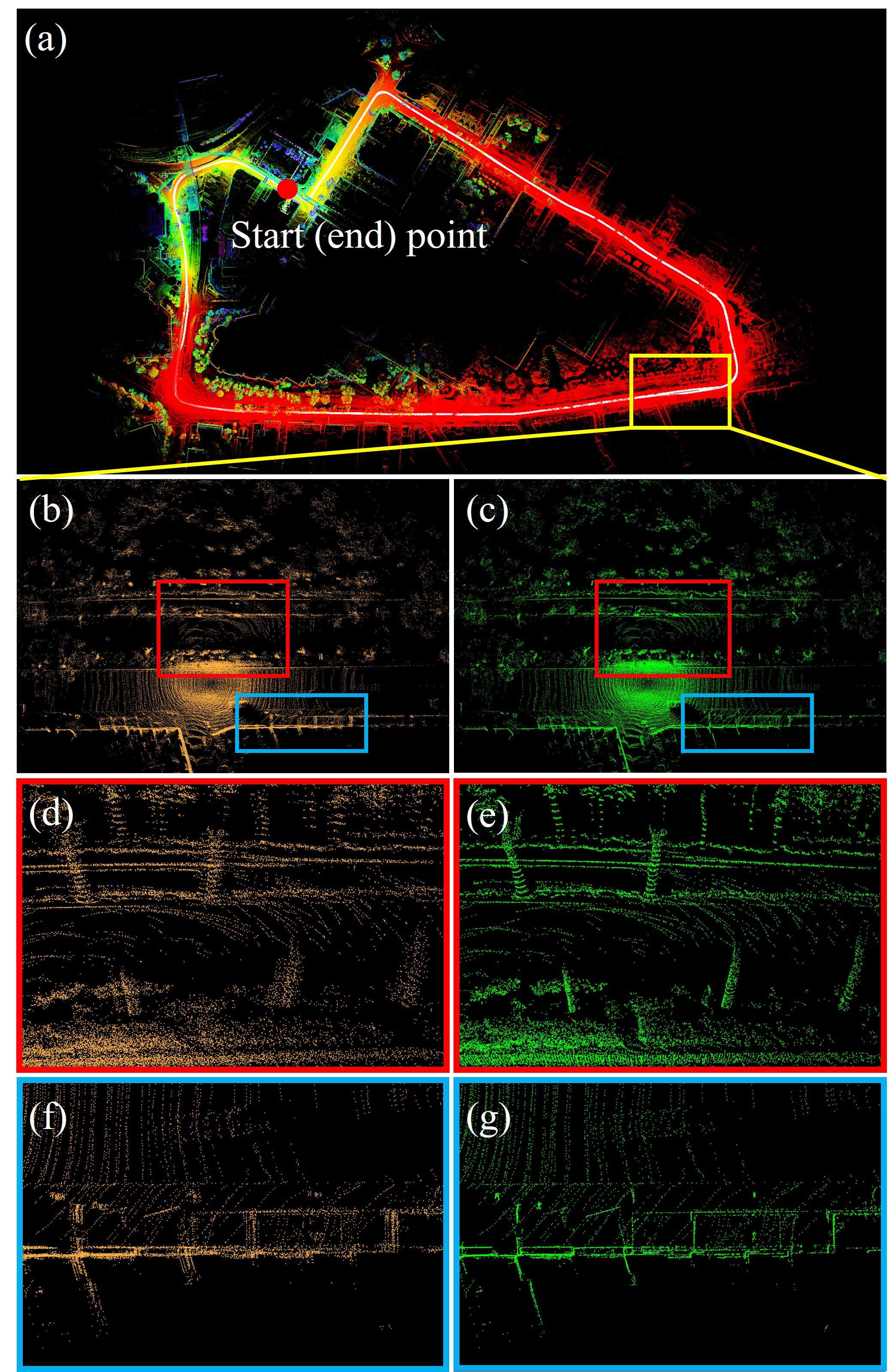}
	\caption{(a) The point cloud map of the sequence ``urban1". (b)-(g) are the point cloud map obtained from the Initialization beginning at the location of yellow frame in (a), which has an initial velocity of 9.6 m/s . The left column of figures (b, d, f) are the map with inaccurate scan states before the BA in initialization, and the right column figures (c, e, g) are the map after the BA in initialization.}
	\label{fig-init_map}
\end{figure}

\subsubsection{Qualitative analysis} \label{sec_initialization_qualitative}


To demonstrate the robustness of our initialization, we tested it on a private sequence called ``private1" collected in an unstructured forest environment. The sequence has fast rotation and quick shaking with an initial angular velocity of $83.1$ deg/s and linear acceleration of $6.7$ m/s$^2$ (excluding the gravity acceleration). Despite the large initial conditions, the Voxel-SLAM can still initialize successfully and generate a trajectory reflecting the real motion with a consistent map, as shown in Fig. \ref{fig-violent-init}(a). The odometry of FAST-LIO2 is divergent at the beginning due to the absence of a robust initialization module. FAST-LIO2 accumulates a short period of IMU measurements to compute an average linear acceleration vector. The vector is then scaled to 9.8 m/s$^2$ to serve as the estimated gravity vector. Since the initial acceleration is up to $6.7$ m/s$^2$ (excluding the gravity acceleration), the direction of the gravity vector computed by FAST-LIO2 is far from the truth, causing the divergence as shown in Fig. \ref{fig-violent-init}(b). To further prove the effectiveness of the initialization module, we start our system at various beginning time along the sequence, ranging from $\frac{1}{7}$ to $\frac{6}{7}$ of its duration. The corresponding results are presented in Fig. \ref{fig-violent-init}(c) to (h), respectively. These results further prove the robustness of the initialization module under various initial conditions and environments.


\subsubsection{Quantitative analysis}

Next, we conduct a quantitative analysis of the initialization using the sequence ``urban1" in UrbanNav \cite{hsu2023hong}. This dataset was collected on a highway and provides the ground-truth poses and velocity at $1$ Hz. The initialization module is executed at different starting time along this sequence. In each case, we evaluate the initial velocity and gravity vector, which are two important parameters affecting the initialization. For the ground-truth required in the evaluation, ``urban1" provides the ground-truth velocity in the IMU body frame. After each initialization, we transform the velocity of the estimated IMU states into their body frame to compare with the ground-truth directly. For the ground-truth gravity vector, it is not provided directly from the dataset, so we compute the average IMU acceleration in the first one second of the sequence as the ground-truth gravity vector. We confirm the IMU is stationary in this period by the ground-truth velocity and observation of the camera images (the IMU and camera are in rigid connection). The relative scan poses and bias computed from the initialization are not evaluated. The reason is that the ground-truth relative scan poses are unavailable due to the low frequency of the ground-truth ($1$ Hz), and the ground-truth IMU bias is also inaccessible in the real world. Besides starting from different time in the sequence, we also explore the impact of different sliding windows (5, 10, 20, and 40) on the performance of the initialization module. 


\begin{figure} [t]
	\centering
	\includegraphics[width=1.0\linewidth]{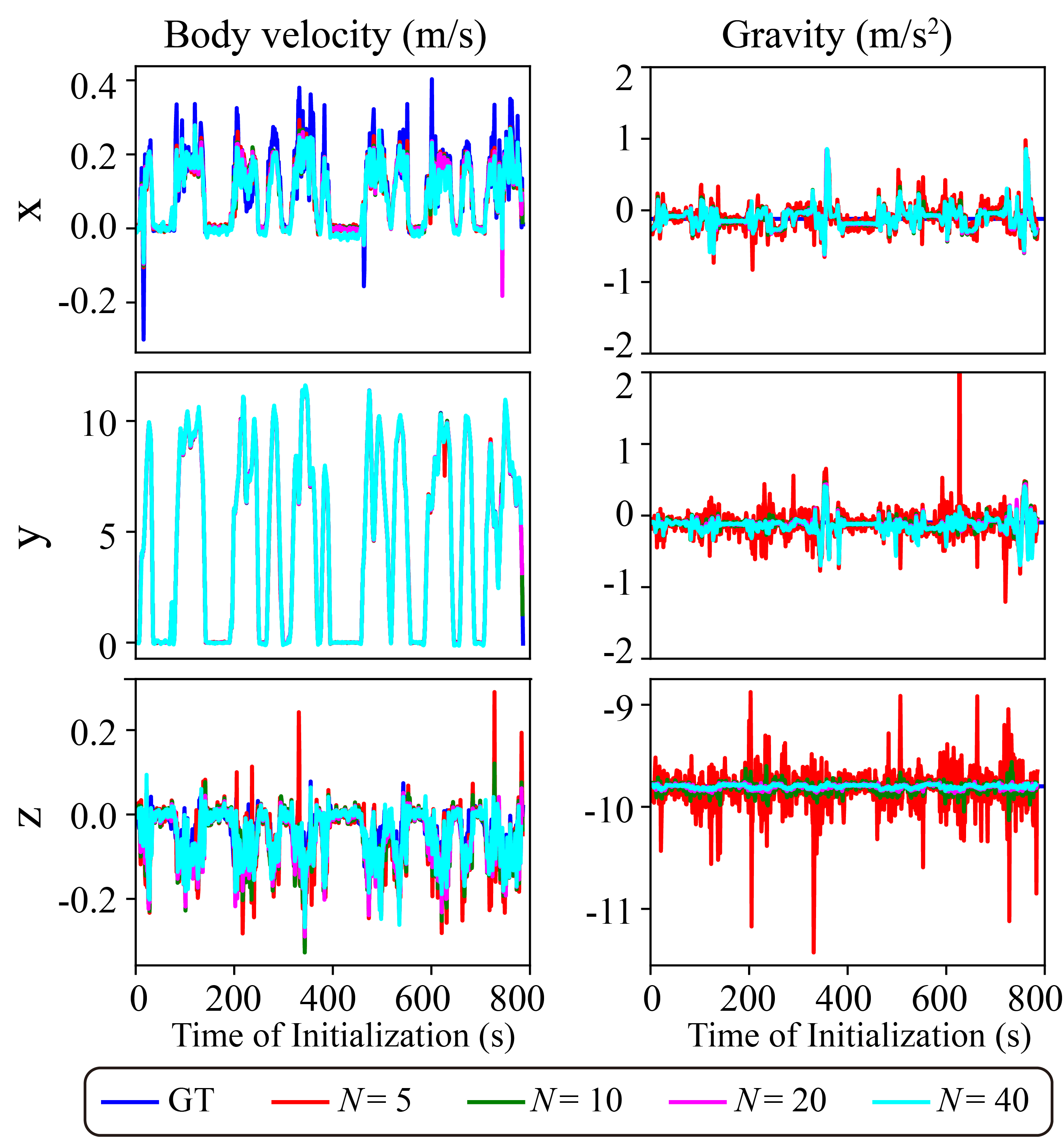}
	\caption{The estimated initial velocity and gravity vector in the sequence ``urban1". The X axis of each subplot is the time along the sequence when the initialization starts executing. Different colors in all plots represent the estimation with different sliding window size.}
	\label{fig-vg_xyz}
\end{figure}

\begin{table}
	\caption{Estimation error (RMSE) and average computation time for different window sizes in the initialization} \label{table-vg-error}
	\centering
	{
		\begin{tabular}{lrrrr}
			\toprule
			Window size ($N$) & 5 & 10 & 20 & 40 \\
			\midrule
			Velocity error (m/s) & 0.1449 & 0.1247 & 0.1239 & 0.1233 \\
			Gravity error (m/s$^2$) & 0.4636 & 0.3057 & 0.2801 & 0.2748 \\
			Computation time (s) & 0.1423 & 0.2910 & 0.6938 & 2.1138 \\
			\bottomrule
		\end{tabular}
	}
\end{table}


The point cloud map from one of the initialization experiments in a mixed structured and unstructured environment is shown in Fig. \ref{fig-init_map}. The initial velocity in this case is $9.6$ m/s. Fig. \ref{fig-init_map}(b, d, f) show the point map obtained by LIO, which serves as the input to the BA optimization in the initialization (Section \ref{sec_initialization}). Fig. \ref{fig-init_map}(c, e, g) show the refined point cloud after the coarse-to-fine voxelization and BA optimization of the initialization. We can see the point clouds on the tree trunks and walls have greatly improved consistency, which proves the effectiveness of the initialization.

The comparison between the ground-truth and estimated initial velocity and gravity vectors is shown in Fig. \ref{fig-vg_xyz} and Table \ref{table-vg-error}. We can see the velocity is well estimated for all sizes of sliding windows ranging from 5 to 40, regardless of the time the initialization starts executing and even at very high initial speeds (i.e., more than 10 m/s). For the gravity vector, the estimation errors decrease with the sliding window size. This is because a larger sliding window size exploits longer IMU data, thereby increasing the information. 
These results suggest that our initialization is well able to initialize the system state and initial local map, even at high initial speeds, given a sufficient sliding window size. Considering that the estimation errors of velocity and gravity vector do not decrease noticeably after the sliding window is larger than 10, but a larger sliding window size dramatically increases the computation time, we fix $N = 10$ in Voxel-SLAM.


\subsection{Single-Session SLAM}

In this section, we compare the proposed system Voxel-SLAM against other state-of-the-art open-source LiDAR-(inertial) odometry, including LINS \cite{qin2020lins}, FAST-LIO2 \cite{xu2022fast}, Faster-LIO \cite{bai2022faster}, Point-LIO \cite{he2023point}, and full SLAM systems with loop closure, including LeGO-LOAM \cite{legoloam2018}, LiLi-OM \cite{li2021towards}, LIO-SAM \cite{shan2020lio}, LTA-OM \cite{zou2024lta} (FAST-LIO2 with loop closure). All the comparative methods use their default parameters unless explicitly stated. When evaluating the accuracy of odometry, we disable the loop closure (LC) for all SLAM methods mentioned above. Specially, we perform ablation studies of our system with odometry (Odom), local mapping (LM), loop closure (LC), and global mapping (GM). Our system with relevant modules is denoted as ``Our (Odom)", ``Our (Odom + LM)", ``Our (Odom + LM + LC)", and ``Our (Full)", respectively.

\subsubsection{Hilti handheld dataset}

\begin{table*}[ht] 
\setlength\tabcolsep{6pt}
	\caption{Absolute trajectory error (RMSE, centimeters) for different odometry and SLAM methods in Hilti} \label{table-single-ate-hilti}
	\centering
	{
		\begin{tabular}{lrrrrrrrrrrrrr}
			\toprule
			Sequences (cm) & hilti01 & hilti02 & hilti03 & hilti04 & hilti05 & hilti06 & hilti07 & hilti08 & hilti09 & hilti10 & hilti11 & hilti12 & hilti13 \\
			\midrule
			\multicolumn{6}{l}{(\textit{odometry without LC})} \\
			LeGO-LOAM & 9.1 & 47 & - & 25.3 & - & 67 & - & 25.3 & 12.7 & 14.3 & 27.1 & 19.7 & -\\
			LiLi-OM & 6.2 & 22.2 & - & 31 & - & 28.9 & - & 20.3 & 6.9 & 8.5 & 19.9 & 25.3 & -\\
			LINS & 6.5 & 18.8 & - & 20.7 & - & 23.1 & - & 17.8 & 7.5 & 9.9 & 28.1 & 20.0 & -\\
			LIO-SAM & 7.4 & 15.2 & - & 23.4 & - & 17.4 & - & 22.4 & 6.6 & 6.8 & 17.6 & 16.8 & 74 \\
			FAST-LIO2 & 1.3 & 2.8 & 32 & 6.7 & 55 & 2.4 & 72 & 1.7 & 2.4 & 1.8 & 4.2 & 3.5 & 16 \\
			Faster-LIO & 1.1 & 2.1 & 37 & 5.0 & 73 & 1.4 & 61 & 2.4 & 1.9 & 2.3 & 2.7 & 2.6 & 11.4\\
			Point-LIO & 1.1 & 3.0 & 23 & 3.7 & 44 & 0.9 & 45 & 2.6 & 3.2 & 1.6 & 3.6 & 4.0 & 9.2\\
			Our (Odom) & 1.3 & 2.5 & 9.3 & 4.2 & 23 & 1.6 & 15.7 & 1.8 & 1.6 & 2.0 & 2.8 & 2.4 & 4.3\\
			Our (Odom+LM) & \textbf{0.8} & \textbf{1.8} & \textbf{3.0} & \textbf{3.4} & \textbf{15.9} & \textbf{0.9} & \textbf{9.8} & \textbf{1.2} & \textbf{1.25} & \textbf{1.4} & \textbf{2.4} & \textbf{1.4} & \textbf{1.26} \\
			\midrule
			\multicolumn{6}{l}{(\textit{full SLAM with LC})} \\
			LeGO-LOAM & 8.8 & 39 & - & 25.3 & - & 67 & - & 22.7 & 12.6 & 12.9 & 27.1 & 16.2 & - \\
			LiLi-OM & 6.2 & 14.2 & - & 31 & - & 27.1 & - & 18.6 & 6.9 & 8.0 & 19.9 & 22.8 & - \\
			LIO-SAM & 6.1 & 10.1 & - & 23.4 & - & 13.4 & - & 17.2 & 6.6 & 5.5 & 17.6 & 12.5 & 74\\
			LTA-OM & 1.27 & 2.5 & 33 & 6.7 & 40 & 2.0 & 65 & 1.2 & 2.4 & 1.78 & 4.1 & 2.9 & 16\\
			Our (Odom+LM+LC) & 0.78 & 1.8 & \textbf{2.8} & 3.4 & 14.2 & 0.9 & 9.2 & 0.90 & 1.25 & 1.4 & 2.4 & 1.4 & 1.26\\
			Our (Full) & \textbf{0.62} & \textbf{1.4} & 2.9 & \textbf{3.3} & \textbf{13.8} & \textbf{0.7} & \textbf{7.8} & \textbf{0.82} & \textbf{1.00} & \textbf{1.14} & \textbf{1.30} & \textbf{1.22} & \textbf{0.80} \\
			\bottomrule
		\end{tabular}
	}
\end{table*}


\begin{figure} [t]
	\centering
	\includegraphics[width=0.95\linewidth]{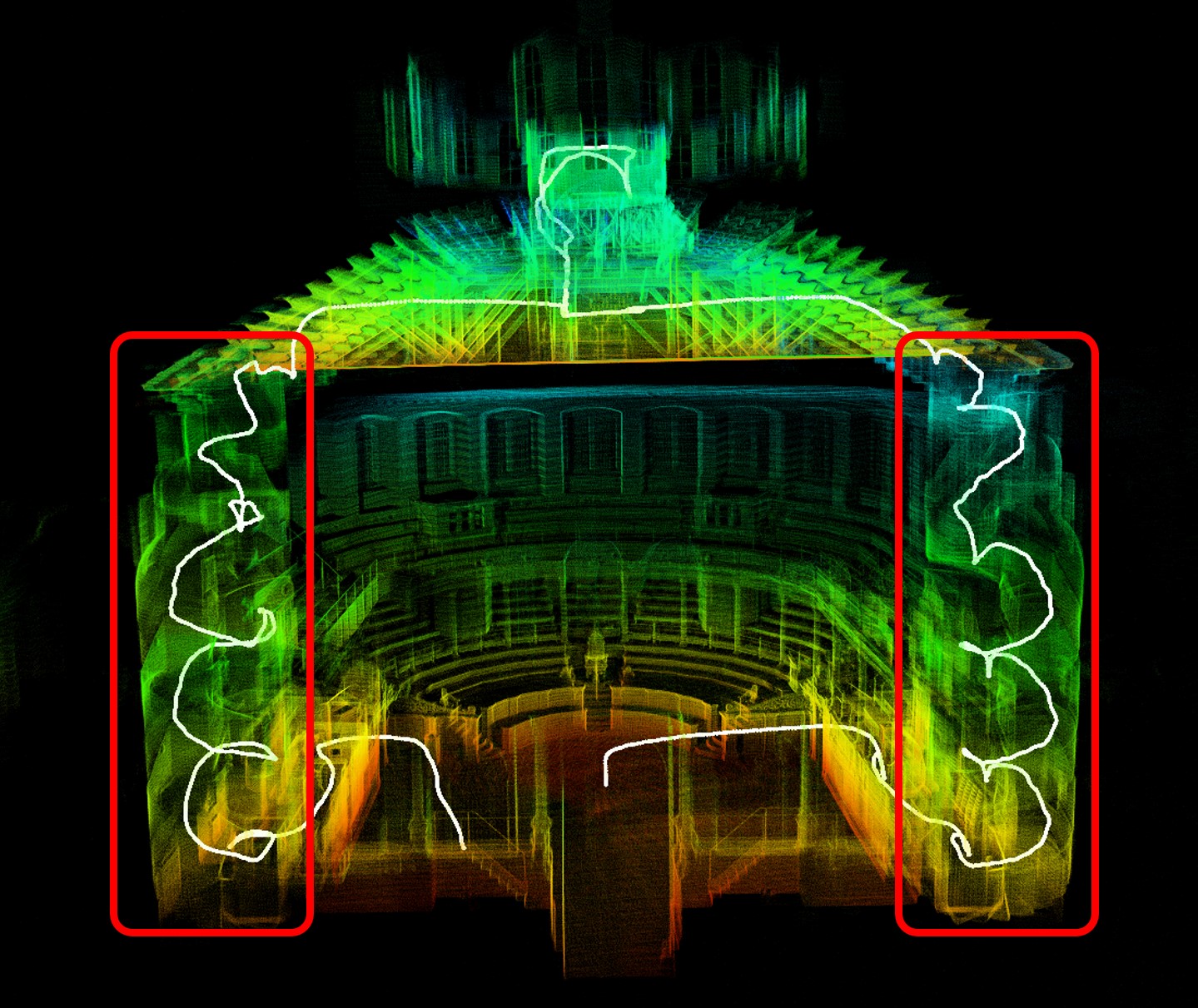}
	\caption{The point cloud map of sequence ``hilti05", which is built by our full Voxel-SLAM system. The red frame shows the trajectory with point cloud map in the narrow staircase.}
	\label{fig_hilti_exp09}
\end{figure}

Table \ref{table-single-ate-hilti} lists the absolute trajectory error (ATE) for all methods in odometry and SLAM. The main difficulty of the Hilti dataset is crossing the narrow staircases, which is a great challenge for LiDAR odometry due to the limited feature constraints and the potential occurrence of degeneration (e.g., sequences 03, 05, 07, and 13), as shown in the red frame of Fig. \ref{fig_hilti_exp09}. The LeGO-LOAM, LiLi-OM, LINS, and LIO-SAM failed in these challenging sequences. The series of ``LIO" methods, including FAST-LIO2, Faster-LIO, and Point-LIO, demonstrated more robustness in handling the staircases but still had unusually large pose errors. Our odometry with a voxel map mitigates this issue by employing adaptive resolution, leading to much smaller pose errors in these scenarios. Additionally, by incorporating local mapping, our odometry improves the accuracy of staircases significantly and achieves the highest accuracy among all sequences. The main reason is the exploitation of a longer data association for estimation compared to the short-term data association in scan-to-map registration used by all previous LIO methods.

The ATE of all SLAM methods with LC are shown in Table \ref{table-single-ate-hilti}. In general, all SLAM methods achieve improved accuracy than their corresponding odometry due to the consideration of additional loop closure constraints. Comparing among all SLAM methods, our method, even without the global mapping, has achieved the highest accuracy among all sequences. With global mapping, our method further reduces all trajectory errors. These accuracy improvements confirm that the global mapping yields improved optimization results across all sequences due to the exploitation of long-term data association. Even in cases where no loop retrieval is detected (e.g., sequences 04, 09, 10, 11, 13), the global mapping can still improve the accuracy, whereas other methods lack a means to further refine their estimates.

\subsubsection{MARS-LVIG aerial mapping dataset}

\begin{table*}
\setlength\tabcolsep{4pt}
	\caption{Absolute trajectory error (RMSE, meters) for different odometry and SLAM methods in MARS-LVIG} \label{table-single-ate-mars}
	\centering
	{
		\begin{tabular}{lrrrrrrrrrrrr}
			\toprule
			Sequences (m) & mars1-1 & mars1-2 & mars1-3 & mars2-1 & mars2-2 & mars2-3 & mars3-1 & mars3-2 & mars3-3 & mars4-1 & mars4-2 & mars4-3\\
			\midrule
			\multicolumn{6}{l}{(\textit{odometry without LC})} \\
			LiLi-OM & 4.56 & 4.81 & 5.35 & 3.68 & 3.72 & 3.70 & 10.54 & 11.66 & 13.08 & - & - & - \\
			LIO-SAM & 3.77 & 4.02 & 4.79 & 1.22 & 1.30 & 1.39 & 6.89 & 6.95 & 8.62 & 12.09 & 11.53 & 14.64 \\
			FAST-LIO2 & 0.66 & 0.46 & 0.48 & 0.26 & 0.39 & 0.59 & 2.17 & 2.05 & 2.51 & 4.46 & 6.54 & 8.37 \\
			Faster-LIO & 0.42 & 0.56 & 0.51 & 0.27 & 0.36 & \textbf{0.32} & 1.71 & 1.49 & 3.12 & 5.50 & 7.43 & 8.79 \\
			Point-LIO & 1.53 & 1.44 & 1.47 & 0.35 & 0.40 & 0.69 & 3.63 & 3.38 & 4.23 & 8.92 & 9.57 & 12.45\\
			Our (Odom) & 0.45 & 0.42 & 0.50 & 0.24 & 0.42 & 0.48 & 2.25 & 1.39 & 1.90 & 4.21 & 5.33 & 7.68\\
			Our (Odom+LM) & \textbf{0.23} & \textbf{0.35} & \textbf{0.32} & \textbf{0.16} & \textbf{0.25} & {0.33} & \textbf{1.18} & \textbf{1.17} & \textbf{1.29} & \textbf{2.49} & \textbf{2.87} & \textbf{3.28} \\
			\midrule
			\multicolumn{6}{l}{(\textit{full SLAM with LC})} \\
			LiLi-OM & 3.83 & 4.04 & 4.10 & 2.87 & 2.93 & 2.85 & 8.47 & 8.86 & 11.55 & - & - & - \\
			LIO-SAM & 2.90 & 2.91 & 3.39 & 0.55 & 0.67 & 0.70 & 6.36 & 5.73 & 7.27 & 12.09 & 11.53 & 14.64 \\
			LTA-OM & 0.49 & 0.42 & 0.39 & 0.22 & 0.31 & 0.40 & 1.09 & 1.34 & 1.47 & 4.46 & 6.54 & 8.37 \\
			Our (Odom+LM+LC) & 0.22 & 0.25 & 0.28 & 0.16 & 0.24 & 0.30 & 0.98 & 0.99 & 1.05 & 2.49 & 2.87 & 3.28 \\
			Our (Full) & \textbf{0.18} & \textbf{0.22} & \textbf{0.26} & \textbf{0.15} & \textbf{0.22} & \textbf{0.29} & \textbf{0.84} & \textbf{0.76} & \textbf{0.70} & \textbf{0.51} & \textbf{0.55} & \textbf{0.77} \\
			\bottomrule
		\end{tabular}
	}
\end{table*}

\begin{figure} [t]
	\centering
	\includegraphics[width=1.0\linewidth]{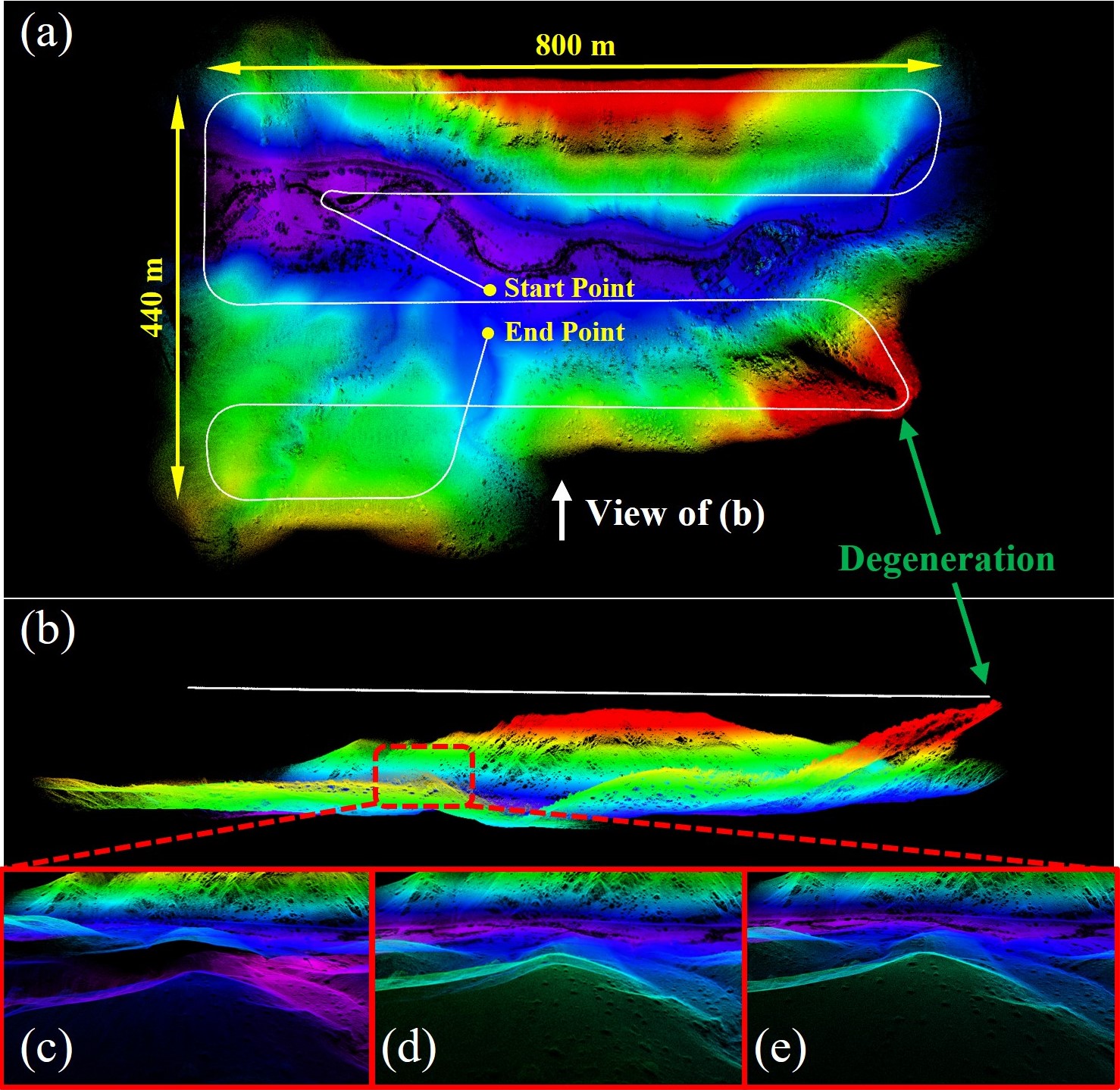}
	\caption{The overview of ``mars4-3". (a)-(b) The global point cloud map in the top view and front view. The point map is built by our full Voxel-SLAM system. (c)-(e) The local map built by FAST-LIO2, our odometry with local mapping, and our full system with global mapping, respectively, for area in the red dashed frame of (a).}
	\label{fig-AMvalley03}
\end{figure}

The MARS-LVIG dataset \cite{li2024mars} has a unique aerial downward-looking viewpoint from about 100 m above the ground. LeGO-LOAM and LINS cannot run this dataset since their special feature extraction is not applicable to the Livox AVIA LiDAR. We follow the instructions from LIO-SAM to adapt it to the LiDAR, and the rest methods can work on Livox AVIA LiDAR naturally without modification. The ATE results of all the methods on this dataset are listed in Table \ref{table-single-ate-mars}. Sequence names sharing the same first number have the same paths but at different flight speeds. Different from the structured and small environment and slow movement in the Hilti dataset, the MARS-LVIG flies fast with the highest speed of 12 m/s in a large scene with massive natural scenes of woods, hills, and rivers, as shown in Fig. \ref{fig-AMvalley03}(a). Our method is still robust in these unstructured environments with fast motion and achieves the best accuracy in odometry and SLAM methods in most sequences.

\begin{table} [t]
\setlength\tabcolsep{4pt}
	\caption{Absolute trajectory error (RMSE, meters) for odometry and SLAM of different methods in UrbanNav} \label{table-single-ate-urbannav}
	\centering
	{
		\begin{tabular}{lrrrrr}
			\toprule
			Sequences (m) & urban1 & urban2 & urban3 & urban4 & urban5 \\
			\midrule
			\multicolumn{6}{l}{(\textit{odometry without LC})} \\
			LeGO-LOAM & 17.93 & 15.94 & 11.87 & 8.53 & 4.92 \\
			LIO-SAM & 10.30 & 11.75 & 5.90 & 5.66 & 1.72 \\
			LiLi-OM & 11.14 & 108.22 & 6.23 & 7.14 & 4.12 \\
			LINS & 12.93 & 75.93 & 17.18 & 9.91 & 5.21 \\
			FAST-LIO2 & 9.08 & 7.20 & 5.22 & 3.11 & 1.22 \\
			Faster-LIO & 9.59 & 8.14 & 5.25 &  3.23 & 1.62 \\
			Point-LIO & 8.14 & 10.53 & 3.46 & 3.75 & 1.18 \\
			Our (Odom) & 7.83 & 9.61 & 4.23 & 3.30 & 1.14 \\
			Our (Odom+LM) & \textbf{5.62} & \textbf{7.13} & \textbf{3.45} & \textbf{2.98} & \textbf{1.04} \\
			\midrule
			\multicolumn{6}{l}{(\textit{full SLAM with LC})} \\
			LeGO-LOAM & 12.54 & 10.68 & 7.23 & 6.43 & 2.87 \\
			LIO-SAM & 8.44 & 8.87 & 3.42 & 3.01 & 1.17 \\
			LiLi-OM & 9.43 & 108.01 & 4.76 & 6.52 & 2.79 \\
			LTA-OM & 5.28 & 5.51 & 2.92 & 2.73 & 0.93 \\
			Our (Odom+LM+LC) & 3.02 & 4.77 & 2.15 & 2.58 & 0.92\\
			Our (Full) & \textbf{2.82} & \textbf{4.25} & \textbf{1.74} & \textbf{2.42} & \textbf{0.90} \\
			\bottomrule
		\end{tabular}
	}
\end{table}

Specially, the fourth group of sequences, ``mars4-1", ``mars4-2", and ``mars4-3", presents the most challenging scenarios. A short period of degeneration occurs when the drone flies to the hilltop shown in Fig. \ref{fig-AMvalley03}(a)-(b). Although most odometry methods except LiLi-OM survived, their trajectories become tilted after the hilltop, with ATEs significantly larger than other sequences in Table \ref{table-single-ate-mars}. Due to the degeneration, the point cloud maps of these odometry methods exhibit inconsistencies, as shown in Fig. \ref{fig-AMvalley03}(c)-(d), where points on the hilltop separate into two layers. As can be seen, the separation by our odometry with local mapping (Fig. \ref{fig-AMvalley03}(d)) is notably smaller than that of the representative LIO method, FAST-LIO2 (Fig. \ref{fig-AMvalley03}(c)), due to the use of mid-term data associations. Accordingly, the quantitative ATEs of our odometry with local mapping, as shown in Table \ref{table-single-ate-mars}, are much better than other LIO methods. For the full SLAM, all the methods fail to identify valid loop detections in these three sequences, leading to ATEs identical to their corresponding odometry. However, with the global mapping, our method can still improve the trajectory accuracy through the coarse-to-fine voxelization and BA optimization (Section \ref{sec_global_mapping}). The coarse-to-fine voxelization voxelizes the separated layers of point clouds into the same voxel, hence associating them into one feature that is optimized in the subsequent BA optimization. The resultant map consistency at this location is greatly improved, as shown in Fig. \ref{fig-AMvalley03}(e). Accordingly, the ATEs of our full method with global mapping in these three sequences are also improved noticeably, as shown in Table \ref{table-single-ate-mars}.


\subsubsection{UrbanNav robotcar urban dataset}

Autonomous driving represents a significant application domain for LiDAR SLAM, and we evaluated our method on the UrbanNav robotcar dataset. The UrbanNav dataset is collected in urban scenarios with a higher speed (maximum speed of 13 m/s), longer collection time (the longest 56 minutes), more dynamic objects, and previously revisited places than the Hilti. The results in Table \ref{table-single-ate-urbannav} demonstrate the effectiveness of our method in urban scenarios, where it achieves the highest accuracy among all sequences and all evaluated methods in both cases of odometry and full SLAM.


\subsection{Multi-Session SLAM}

\begin{figure} [t]
	\centering
	\includegraphics[width=1.0\linewidth]{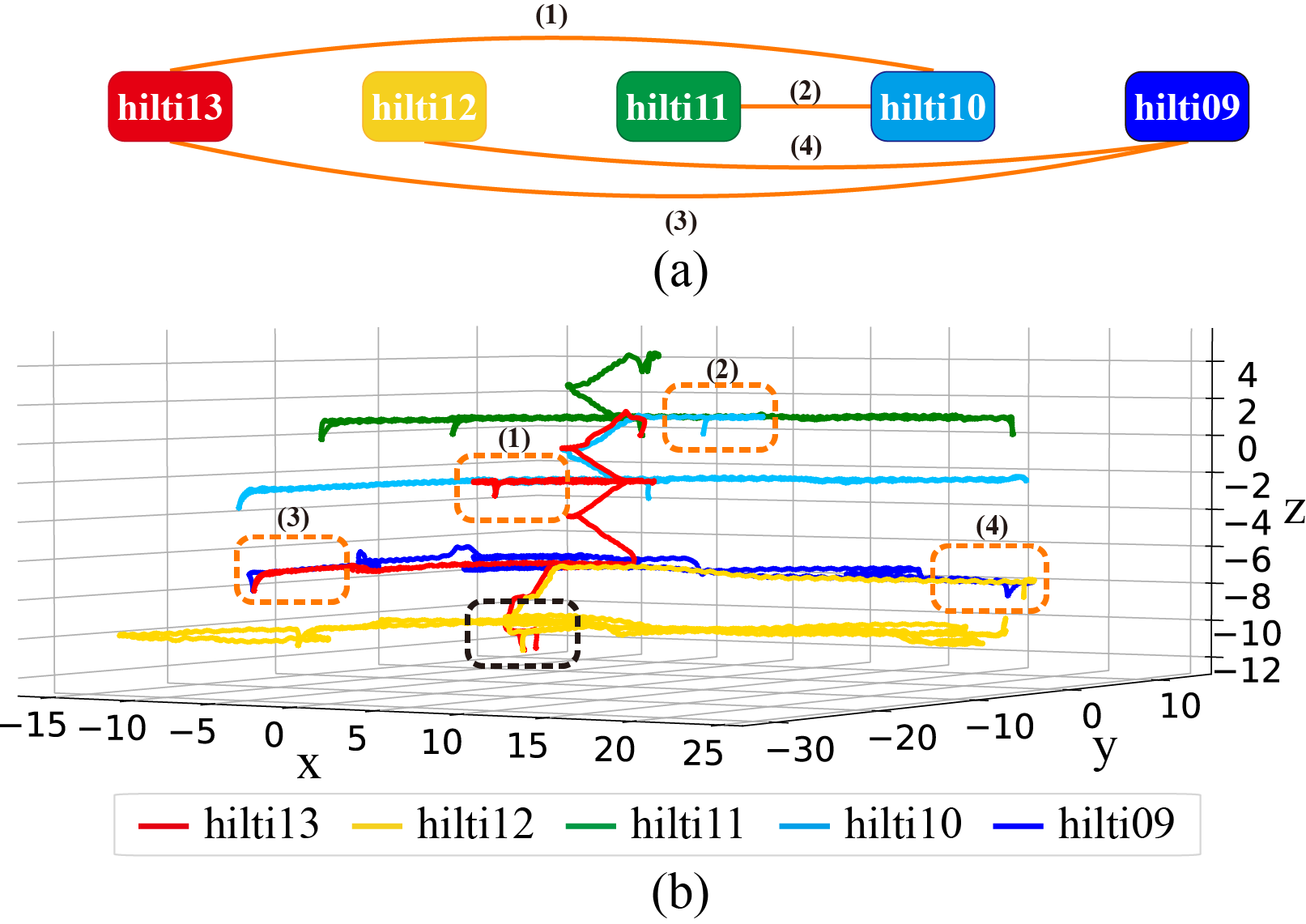}
	\caption{(a) The connectivity among the 5 sequences of Hliti 09 - 13. The numbers on the connected edges represent the order of the detected loops. (b) The trajectories of the 5 sequences, estimated by the full Voxel-SLAM with global mapping. The orange dashed frames show detected loops in (a). The black dashed frame shows a place overlapped by sequence ``hilti13" and ``hilti12" but is not detected as a loop due to the missing detection. }
	\label{fig-multi_connect}
\end{figure}

The sequences from Hilti 09-13 are collected at the same construction site, and their ground truth shares the common world frame. We use these sequences to test the multi-session localization ability of our Voxel-SLAM. Firstly, Voxel-SLAM is fed with the sequences ``hilti13", ``hilti12", ``hilti11", ``hilti10", and ``hilti09", sequentially. After completing each sequence, Voxel-SLAM saves it in memory as a previous session, enabling the system to retrieve loop detections among these previous sessions in future sequences. Running through all five sequences, the Voxel-SLAM builds a connectivity graph as shown in Fig. \ref{fig-multi_connect}(a), which successfully detects loops at revisited areas among different sequences, such as the areas labeled as (1-4) in the figure. This leads to a connected pose graph, from which the poses of all five sequences can be optimized jointly by PGO and then by global mapping. The estimated pose trajectory from Voxel-SLAM is shown in Fig. 11(b).

\begin{figure} [t]
	\centering
	\includegraphics[width=0.98\linewidth]{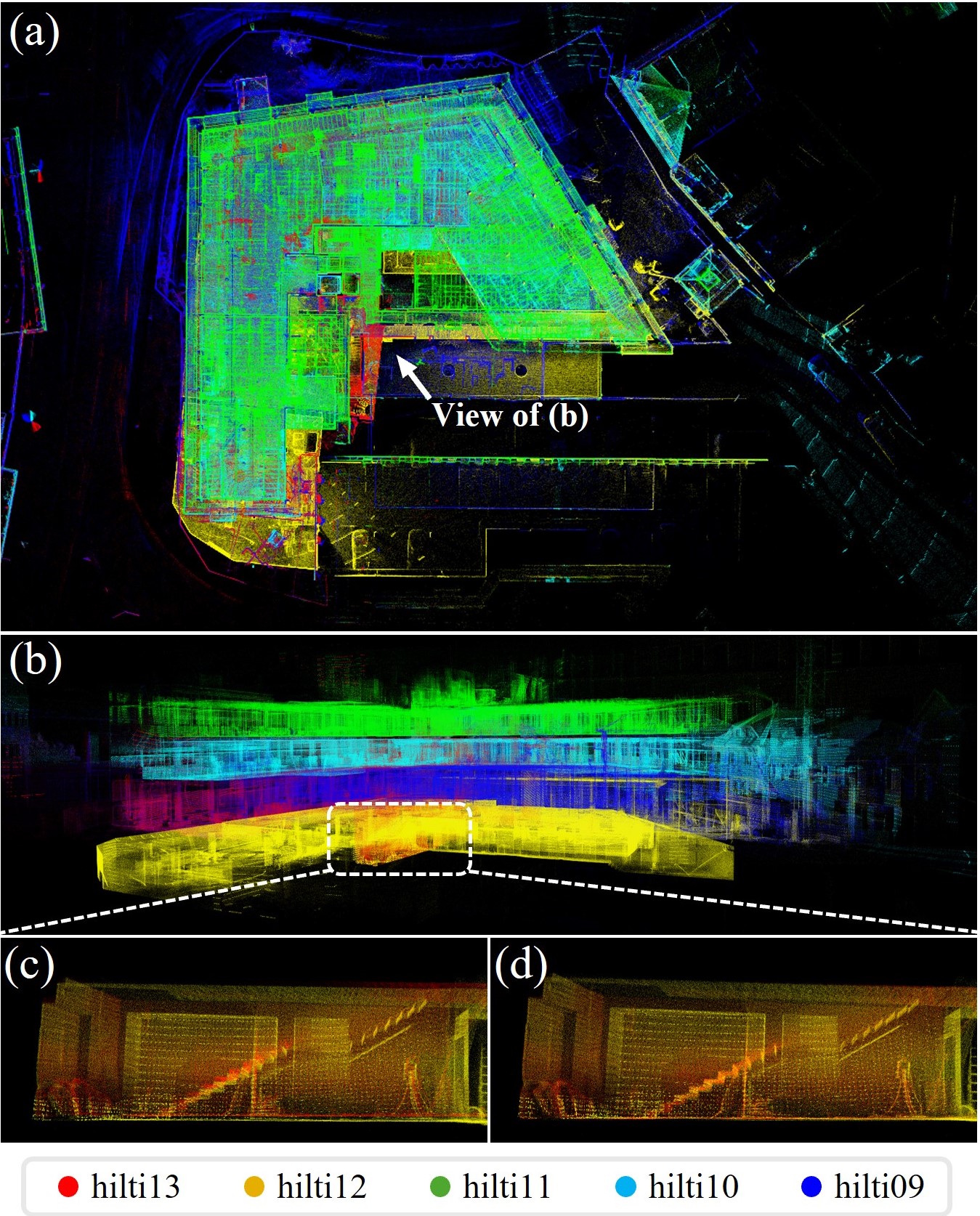}
	\caption{(a)-(b) The global point cloud map of the five sessions from the top view and side view. (c)-(d) The zoomed-in point cloud of the area in the white dashed frame of (b), which is also area in the black dashed frame of Fig. \ref{fig-multi_connect}(b), obtained from PGO and global mapping, respectively.}
	\label{fig-multi_total}
\end{figure}

\begin{table} [t]
\setlength\tabcolsep{3pt}
	\caption{ATE (RMSE, centimeters) for each individual session and full multi-session, estimated by pose graph optimization (PGO) only or with additional global mapping} \label{table-multi-hilti}
	\centering
	{
		\begin{tabular}{lcccccc}
			\toprule
			& hilti09 & hilti10 & hilti11 & hilti12 & hilti13 & Multi-session \\
			\midrule
			PGO only & {1.00} & {1.14} & {1.30} & {1.22} & \textbf{0.80} & 7.6 \\
			Global mapping & \textbf{0.60} & \textbf{0.91} & \textbf{1.13} & \textbf{1.16} & \textbf{0.80} & \textbf{4.9} \\
			\bottomrule
		\end{tabular}
	}
\end{table}

Despite the accuracy and robustness of the BTC descriptors \cite{yuan2024btc} used in our Voxel-SLAM, there could still be missing loop detection (e.g., the black dashed frame in Fig. \ref{fig-multi_connect}(b), which is also marked in the white dashed frame of Fig. \ref{fig-multi_total}(b)). This missing detection occurred in a narrow corridor, which is very challenging for the LiDAR-based place recognition methods, causing insufficient constraints between the ``hilti13" and ``hilti12" and hence limiting the accuracy achievable by the PGO. The corresponding point cloud in this area also exhibits clear inconsistency, as shown in Fig. \ref{fig-multi_total}(c). This issue is effectively addressed by global mapping, as shown in Fig. \ref{fig-multi_total}(d). The entire point cloud maps refined by global mapping are shown in Fig. \ref{fig-multi_total}(a, b), colored by different sequences. Table \ref{table-multi-hilti} shows the ATEs of the trajectories estimated by Voxel-SLAM with PGO only or with additional global mapping. The application of global mapping not only enhances the multi-session accuracy of all five sequences but also improves the accuracy for each individual session, due to the full exploitation of multi-map data association.

The above multi-session experiment further proves the performance enhancements, especially in the map consistencies, brought by global mapping. In the above analysis, we did not compare the results of LTA-OM \cite{zou2024lta}. Although LTA-OM has multi-session SLAM capability, it only supports finding loop closures in one previous session and cannot handle many previous sessions simultaneously. In addition, it fails to find the loop closure between ``hilti11" and ``hilti10".

\subsection{Relocalization}

\begin{figure} [t]
	\centering
	\includegraphics[width=0.98\linewidth]{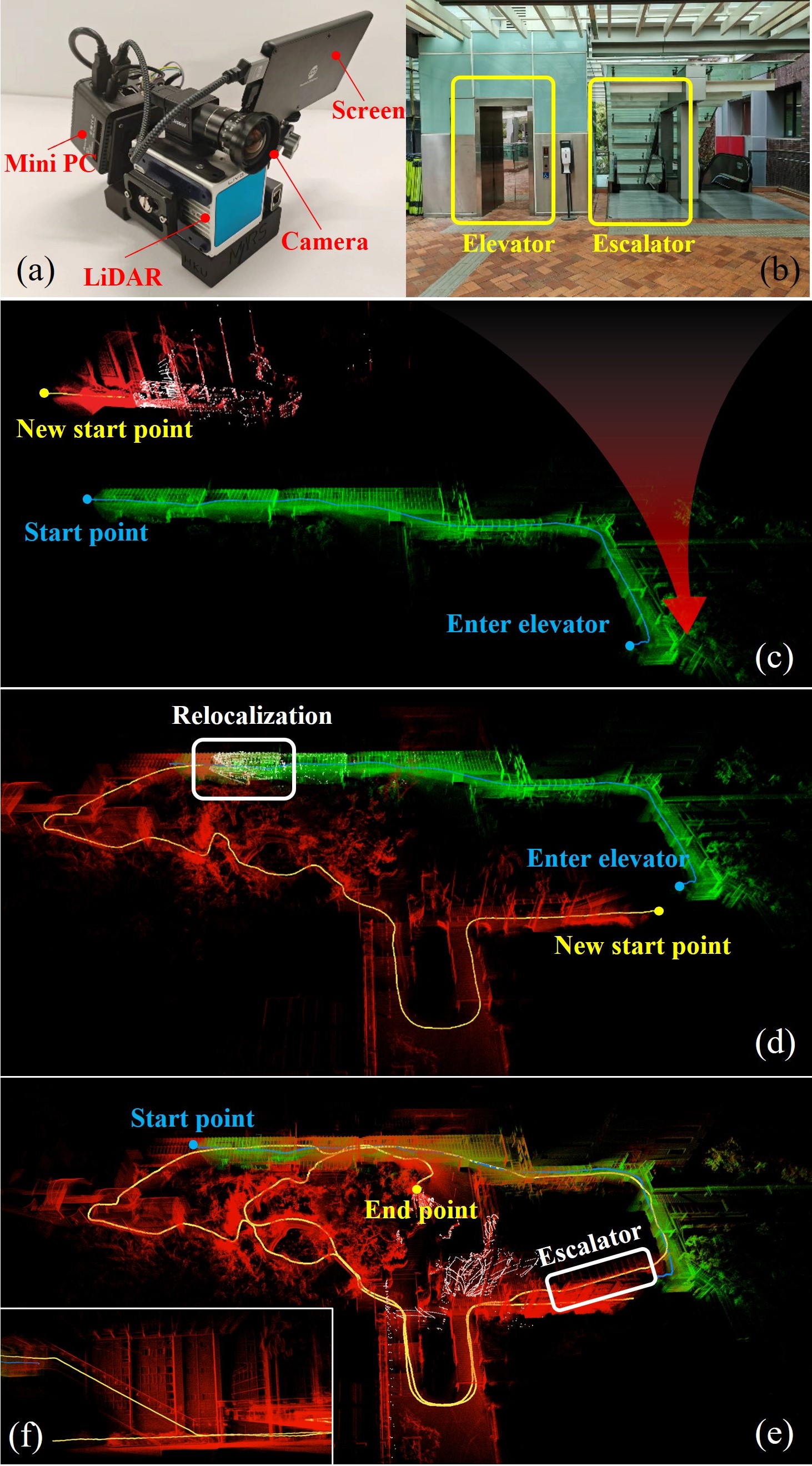}
	\caption{The sequence ``private2". (a) The handheld equipment. (b) The elevator and escalator travelled by the sequence. The elevator will cause a system divergence due to the confined spaces and inconsistent measurements between LiDAR and IMU. (c) The maps and trajectories after detecting divergence in the first session (the green one) and successful re-initialization of the second session (the red one). (d) The maps and trajectories after relocalization of the second session (red) in the first session (green). (e) The maps and trajectories at the end of data collection. (f) The zoomed-in map of the escalator.}
	\label{fig-relc_equipment}
\end{figure}

\begin{figure} [t]
	\centering
	\includegraphics[width=0.98\linewidth]{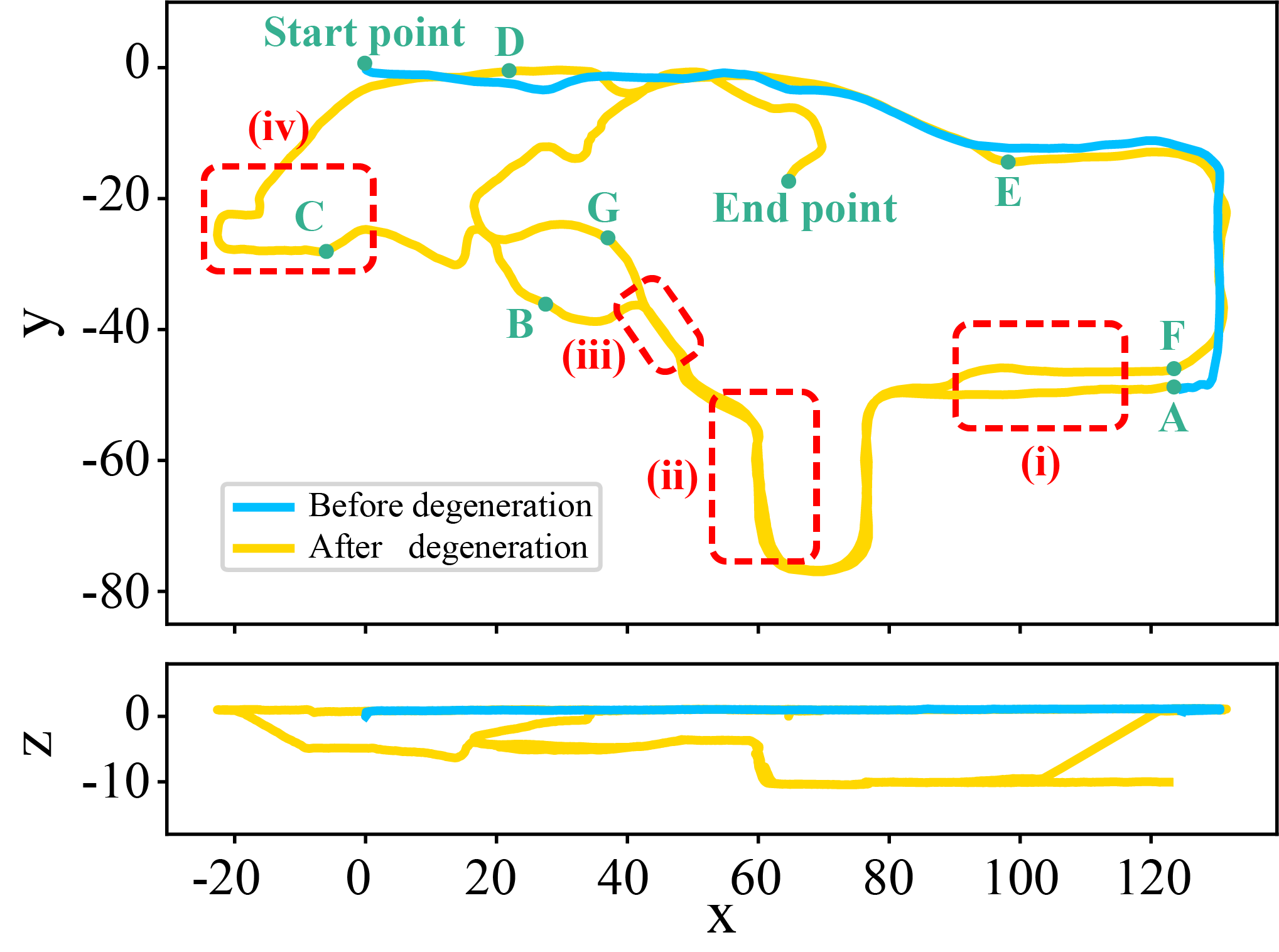}
	\caption{The complete trajectories of data collection from top view and front view. We mark some waypoints from A to G for readers understanding the trajectories better. The sequence begin at start point and walk in the order of the waypoints until to the end point. The red dashed frames are the places whose point clouds will be displayed in Fig. \ref{fig-relc_total}.}
	\label{fig-relc_trajectory}
\end{figure}

\begin{figure} [t]
	\centering
	\includegraphics[width=0.98\linewidth]{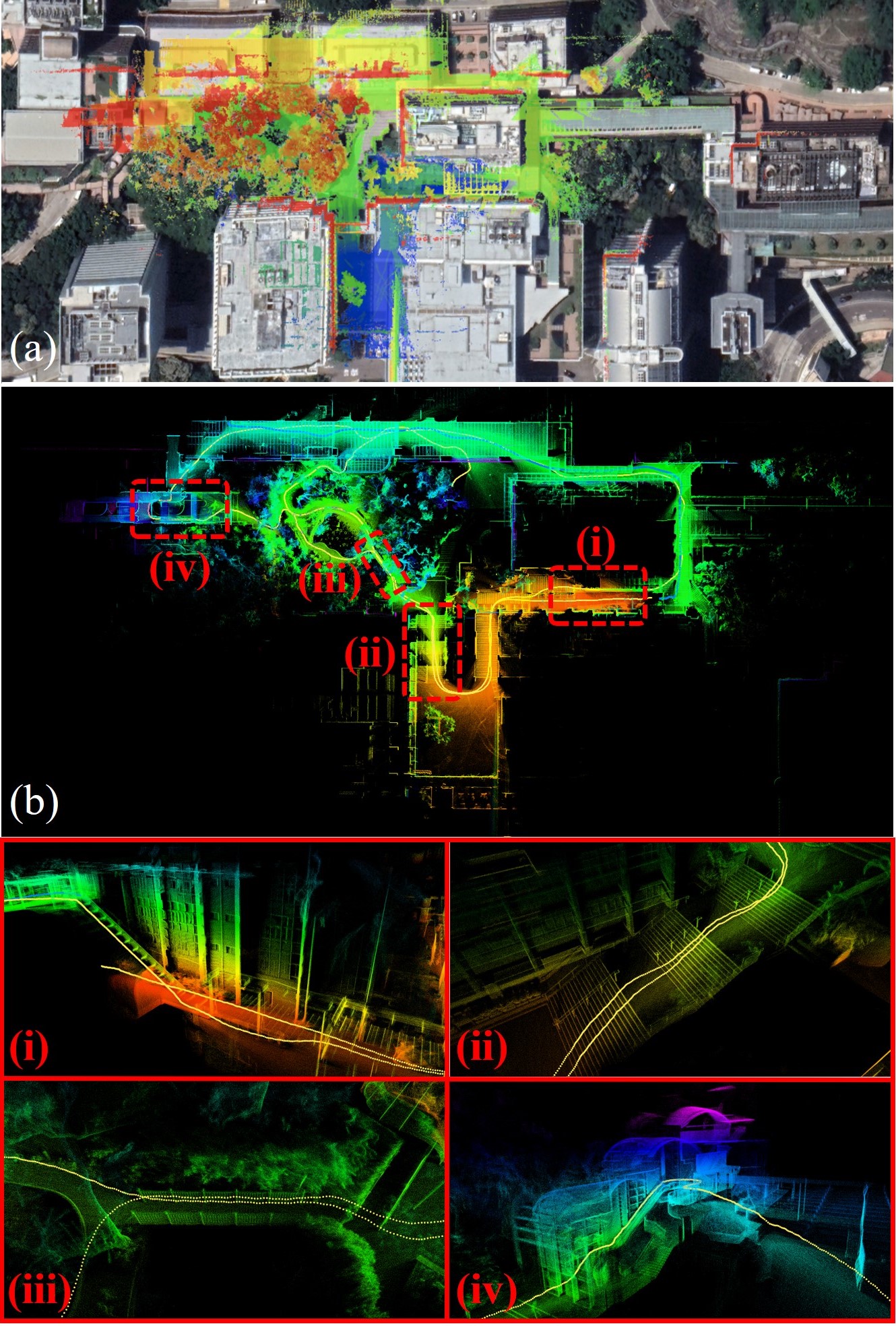}
	\caption{(a) The map aligned with Google Earth. (b) The point cloud map optimized by our system. (i)-(iv) The zoomed-in maps from the red dashed frames in (b).}
	\label{fig-relc_total}
\end{figure}

This experiment evaluates the relocalization capability of the system after encountering a degeneration. Moreover, considering the application in mobile robots with limited computation resources, we evaluate the system on the onboard computer equipped with an Intel i3-N305 CPU. The used sequence ``private2" was collected using the handheld device. The experiment tests all functions of Voxel-SLAM, encompassing initialization (and re-initialization) in the middle of data collection with non-zero state, odometry, local mapping, loop closure detection across multiple sessions, relocalization, and global mapping. All of these functionalities were executed online on the onboard computer.

As shown in Fig. \ref{fig-relc_equipment}, the sequence is collected in a campus environment with many moving pedestrians, significant variations in height, confined spaces (e.g., inside elevators) causing degeneration and motion ambiguity, and a mix of structured and unstructured surroundings. The collecting trajectory is shown in Fig. \ref{fig-relc_trajectory} from start to end in order of the waypoints A-G. Specifically, after taking off from the start point, we walk along the blue path, which renders the green point cloud in Fig. \ref{fig-relc_equipment}(c), until entering the elevator (waypoint A), as shown in Fig. \ref{fig-relc_equipment}(b). Due to the confined space in the elevator and the inconsistency between LiDAR and IMU measurements, Voxel-SLAM detects the system divergence and starts a new session. The new session initialization keeps running but fails until we exit the elevator, where the successfully initialized new local map is shown as the red point cloud in Fig. \ref{fig-relc_equipment}(c). After exiting the elevator, we pass a small park with woods and vegetation (waypoint B) and a narrow staircase (waypoint C), arriving at waypoint D near the start point of the first session. Revisiting the start point causes Voxel-SLAM to detect loops with the first session and a subsequent relocalization that connects the second session to the first session, as shown in Fig. \ref{fig-relc_equipment}(d). We continue to walk along the previous path (passing through waypoint E), where long-term associations with the first session are consistently exploited by Voxel-SLAM due to its loop closures and keyframe loading in local mapping (Section \ref{sec_voxelization}). Then, we return to the elevator (waypoint F) that caused the system divergence in the first session, but this time we choose the escalator on the right (Fig. \ref{fig-relc_equipment}(b)). Traveling on the escalator does not cause a degeneration, so the current session proceeds smoothly, as shown in Fig. \ref{fig-relc_equipment}(e). After several loop closures in the current and previous sessions (F $\rightarrow$ G $\rightarrow$ end point), the data collection ends, which triggers the global mapping to optimize the entire point cloud map.

The online relocalization experiment proves the robustness and reliability of Voxel-SLAM. When entering the elevator, the odometry detects the system divergence and restarts the system. The initialization re-initializes constantly in the elevator until we exit the elevator. When returning to previously visited places, Voxel-SLAM can relocalize the current session into the world frame of the previous session. After the relocalization, the long-term and multi-map data associations, including the keyframe loading and loop closure among multiple sessions, are fully exploited to ensure map consistency. The complete trajectories of two sessions are shown in Fig. \ref{fig-relc_trajectory}. As can be seen, the two trajectories are consistent on $z$-axis. The global point cloud map of the whole sequence is shown in Fig. \ref{fig-relc_total} (b), and the map is aligned well with Google Earth\footnote{\url{https://earth.google.com/}} shown in Fig. \ref{fig-relc_total}(a). Fig. \ref{fig-relc_total}(i-iv) shows the point cloud maps from the multiple revisited places by one or multiple sessions. The points across different sessions are consistent due to the full exploitation of long-term and multi-map data associations in the local mapping, loop closure, and global mapping modules.


\subsection{Computation Time}

For all the above datasets and experiments, we evaluate the average time and total memory consumption. Note that the LiDAR scans in all the sequences are all at 10 Hz. To examine the real-time capability of our system, we compute the average computation time of each recurrent modules (odometry, local mapping, loop closure, and keyframe BA) distributed to each scan. For the odometry, and local mapping, they naturally run at the LiDAR scan rate (i.e., 10 Hz), so the time is their actual computation time. For loop closure and keyframe BA, they are triggered by events (e.g., the selection of a key frame or detection of a loop frame), so we accumulate their total computation time over the whole sequence and divide the accumulated time by the number of LiDAR scans to obtain the per scan average time. Since the odometry \& local mapping, loop closure, and keyframe BA are executed in three parallel threads, the average time consumption of each thread should be smaller than the time interval between two consecutive LiDAR scans ($t_s = 0.1$s) to ensure the real-time operation. 

Table \ref{time-mem-cost} shows the time and memory consumption of all sequences in the previous experiments. As can be seen, the time consumption of the three working threads, individually or even in total, is way smaller than the scan interval $t_s$, proving the real-time performance of the Voxel-SLAM even on the onboard computer with constrained computation resources. For the initialization and global mapping modules in Voxel-SLAM, they are generally executed only once prior to or after each session (or sequence), so we evaluate their average time consumption per running. As shown in Table \ref{time-mem-cost}, the time consumption of global mapping after session end is much less than (ranging from 0.6\% to 4.7\%) the sequence duration (i.e., data collection time), proving the efficiency of the designed global mapping module. Specially, in the longest sequence, ``urban3", the global mapping after session end takes 79.7 seconds, versus the data collection duration of 56 minutes and trajectory of 4.86 kilometers.

Memory usage is another important metric for a system exploiting long-term data association, since the system should store all the necessary information in memory to retrieve the long-term data association. As shown in Table \ref{time-mem-cost}, the memory usage is way below the physical memory, even on the onboard computer with constrained resources.

\begin{table*} [t]
\caption{The mean and standard error of time and memory consumption for all sequences in the experiments}
\label{time-mem-cost}
\centering
\begin{threeparttable}
{
\begin{tabular}{lcccc}
\toprule
                                       & Hilti         & MARS-LVIG     & UrbanNav      & Private2 (onboard) \\
\midrule
\textit{(Per scan)}\tnote{\#}           &               &               &                &                    \\
Odometry (sec)                         & 0.008 / 0.003 & 0.023 / 0.007 & 0.013 / 0.002 & 0.008 / 0.002     \\
Local mapping (sec)                    & 0.013 / 0.004 & 0.032 / 0.010 & 0.024 / 0.005 & 0.019 / 0.008     \\
Loop closure (sec)                     & 0.004 / 0.002 & 0.031 / 0.016 & 0.010 / 0.002 & 0.007 / 0.000\tnote{*}      \\
Keyframe BA (sec)                      & 0.002 / 0.001 & 0.009 / 0.003 & 0.004 / 0.001 & 0.004 / 0.000\tnote{*}      \\
Total (sec)                            & 0.027 / 0.009 & 0.095 / 0.028 & 0.051 / 0.006 & 0.037 / 0.000\tnote{*}      \\
\midrule
\textit{(Per running)}\tnote{\#}        &               &               &               &                    \\
Initialization (sec)                   & 0.087 / 0.040 & 0.453 / 0.102 & 0.207 / 0.023 & 0.148 / 0.019      \\
Global mapping after session end (sec) & 3.157 / 2.895 & 20.74 / 16.36 & 34.20 / 26.24 & {8.819 / 8.124}     \\
\midrule
Sequence duration (sec)                & 238.6 / 100.9 & 556.9 / 290.2 & 1295\ \,/\,\ 1118 & 694.0 / 0.000\tnote{*}      \\
Total memory usage (GB)                            & 1.068 / 0.510 & 5.203 / 2.553 & 4.535 / 1.830 & 2.932 / 0.000\tnote{*}      \\
\bottomrule
\end{tabular}
}
\begin{tablenotes}
\footnotesize
\item[*] The standard error is zero since only one sequence is evaluated.
\item[\#] The average time consumption per scan and per running means distributing the total time consumption to each scan and each running, respectively.
\end{tablenotes}
\end{threeparttable}
\end{table*}

\section{Conclusion and Future Works} \label{sec conclusion}

In this paper, we presented the Voxel-SLAM: a complete, accurate, and versatile LiDAR-inertial SLAM system, including the modules of initialization, odometry, local mapping, loop closure, and global mapping, all of which employ the same adaptive voxel map structure. The initialization exhibits remarkable speed and robustness, providing accurate states and a consistent map for the following modules. The odometry swiftly estimates the current state while perceiving potential system divergence. The local mapping utilizes an efficient, tightly-coupled LiDAR-inertial BA to refine the states and map simultaneously, thereby enhancing accuracy and robustness. The loop closure can detect revisited places in multiple sessions. An efficient and accurate global mapping is devised with a data pyramid. Collectively, these modules make full use of four types of data association: short-term, middle-term, long-term, and multi-map.

The system could be extended to fuse image measurements, which can bolster robustness in some degeneration environments, provide color information for point clouds, and enhance place recognition performance. Moreover, although the current system is efficient enough to run in real-time on an onboard computer, GPU parallelization can further accelerate the system efficiency, especially for global mapping. The framework of voxel map is very suitable for parallel operations.

\section{Acknowledgement}

The authors gratefully acknowledge DJI and Livox Technology for fund and equipment support during the development. 

\appendix
\subsection{Iterative Optimization for LiDAR-Inertial BA} \label{iterative BA}
The cost function of the LiDAR-inertial BA is,
\begin{align}
&\arg\min_{\mathcal X} \ c(\mathcal X)  \notag \\
=&\arg\min_{\mathcal X} \left(
\frac{1}{2} \sum_{i=1}^{N-1} \left\| 
\mathbf r_{i,i+1}(\mathcal X)
\right\|^2_{\boldsymbol\Sigma_{i, i+1}^{-1}}
+
\sum_{j=1}^{M} \lambda_{j}^{\text{min}} (\mathcal X)
\right)
\end{align}
where the definition of each item follows (\ref{eq-liba-cost}). Given a perturbation $\delta \mathcal X = [\delta \mathbf x_1, \delta \mathbf x_2,\cdots,\delta \mathbf x_{N}, {\delta \mathbf g}]$ to state $\mathcal X$,
\begin{align}
&c(\mathcal X \boxplus \delta\mathcal X) = 
\notag \\
&\frac{1}{2} \sum_{i=1}^{N-1} \left\| 
\mathbf r_{i,i+1}(\mathcal X \boxplus \delta\mathcal X )
\right\|^2_{\boldsymbol\Sigma_{i, i+1}^{-1}} 
+
\sum_{j=1}^{M} \lambda_{j}^{\text{min}} (\mathcal X \boxplus \delta\mathcal X)
\end{align}

Expand the residual to the highest second-order approximation,

\begin{align}
&c(\mathcal X \boxplus \delta\mathcal X) \approx
\sum_{j=1}^{M} \big(\lambda_{j}^{\text{min}} (\mathcal X) + \mathbf g_j \delta\mathcal X + \frac{1}{2} \delta\mathcal X^T \mathbf H_{j} \delta\mathcal X \big) \notag 
\\
& + \frac{1}{2} \Bigg(\sum_{i=1}^{N-1} \Big(
\left\| \mathbf r_{i,i+1}(\mathcal X) \right\|^2_{\boldsymbol\Sigma_{i,i+1}^{-1}} 
+ 2\mathbf r_{i,i+1}(\mathcal X)^T \boldsymbol\Sigma_{i,i+1}^{-1} \mathbf J_i  \delta\mathcal X  \notag
\\
& + \delta\mathcal X^T \mathbf J_i^T \boldsymbol\Sigma_{i,i+1}^{-1} \mathbf J_i \delta\mathcal X
\Big)\Bigg)
\\
& = \sum_{j=1}^{M} \lambda_{j}^{\text{min}} (\mathcal X) + \frac{1}{2} \sum_{i=1}^{N-1} \left\| \mathbf r_{i,i+1}(\mathcal X) \right\|^2_{\boldsymbol{\Sigma}_{i,i+1}^{-1}} \notag \\
& + \Bigg(\sum_{j=1}^{M} \mathbf g_j + \sum_{i=1}^{N-1} \mathbf r_{i,i+1}(\mathcal X)^T \boldsymbol\Sigma_{i, i+1}^{-1} \mathbf J_i \Bigg) \delta\mathcal X \notag \\
& + \frac{1}{2} \delta\mathcal X^T \Bigg(\sum_{j=1}^{M} \mathbf H_{j} + \sum_{i=1}^{N-1} \mathbf J_i^T \boldsymbol\Sigma_{i, i+1}^{-1} \mathbf J_i \Bigg) \delta\mathcal X \label{quadratic function}
\end{align}
where $\mathbf g_j$ and $\mathbf H_{j}$ are the gradient and Hessian matrix of LiDAR BA factor associated to the $j$-th plane feature, with exact form drawn from \cite{liu2023efficient}. $\mathbf J_i$ is the Jacobian of IMU preintegration residual $\mathbf r_{i,i+1}$, which consists of the derivative of $\mathbf r_{i,i+1}$ with respect to the state $\delta \mathbf x_i$ and that to the gravity vector $\delta \mathbf g$. The former is identical to \cite{forster2016manifold} and the latter is:
\begin{align}
&\frac{\partial \mathbf r_{\Delta \mathbf R_{ij}}}{\partial \delta \mathbf g} = \mathbf 0_{3\times3} \quad\quad\quad\ \ 
\frac{\partial \mathbf r_{\Delta \mathbf p_{ij}}}{\partial \delta \mathbf g} = -\frac{1}{2}\mathbf R_i^T \Delta t_{ij}^2 
\notag 
\\
&\frac{\partial \mathbf r_{\Delta \mathbf v_{ij}}}{\partial \delta \mathbf g} = -\mathbf R_i^T \Delta t_{ij} \qquad
\frac{\partial \mathbf r_{\Delta \mathbf b_{ij}^g}}{\partial \delta \mathbf g} =
\frac{\partial \mathbf r_{\Delta \mathbf b_{ij}^a}}{\partial \delta \mathbf g} = \mathbf 0_{3\times3} \label{eq_g_jacobian}
\end{align}

The quadratic function (\ref{quadratic function}) attains its minimum when 
\begin{align}
\Bigg(\sum_{j=1}^{M} \mathbf H_{j} + & \sum_{i=1}^{N-1} \mathbf J_i^T \boldsymbol\Sigma_{i, i+1}^{-1} \mathbf J_i \Bigg) \delta\mathcal X^{*} = \notag
\\
&-\Bigg(\sum_{j=1}^{M} \mathbf g_j^T + \sum_{i=1}^{N-1} \mathbf J_i^T \boldsymbol\Sigma_{i, i+1}^{-1} \mathbf r_{i,i+1}(\mathcal X) \Bigg)
\end{align}
which leads to the optimum solution $\delta\mathcal X^{*}$. In practice, we optimize the states iteratively within the LM framework until converging.

\subsection{{Criterion of Degeneration}} \label{degeneration judge}




The degeneration of a scene is determined by the distribution of independent plane constraints contained in the scene. If there are plane constraints in all three directions, the scene is considered non-degenerated. Otherwise, the scene is degenerating. In practice, the directional distribution of the plane features contained in a scene can be examined from the eigenvalues of the matrix,
\begin{align}
	\mathbf M = \sum \mathbf n_i \mathbf n_i^T
\end{align}
where $\mathbf M\in\mathbb R^{3\times3}$ and $\mathbf n_i \in \mathbb S^2$ is the normal vector of the $i$-th plane feature in the scene. The three eigenvalues of the matrix $\mathbf M$ indicate the sufficiency of the constraints along each eigenvector direction. Consequently, we only need to look at the minimum eigenvalue: if the minimum eigenvalue is larger than the predetermined threshold, it signifies that there are enough plane constraints even in the most insufficient direction, indicating the scene is not degenerate; otherwise, the scene is considered degenerated.

\subsection{Detail information of all sequences} \label{appen_sequence_list}

The detail information about total 32 sequences tested in Section \ref{sec experiments} are listed in Table \ref{sequence-information}.

\begin{table}
\caption{Details of all the sequences for experiments}
\label{sequence-information}
\centering
{
\begin{tabular}{llrr}
\toprule
Sequence & Name & Duration & Distance \\
& & (min:sec) & (km) \\
\midrule
hilti01 & exp01-construction & 3:47 & 0.16 \\
hilti02 & exp02-construction & 7:10 & 0.31 \\
hilti03 & exp03-construction & 5:09 & 0.23 \\
hilti04 & exp07-long-corridor & 2:12 & 0.11 \\
hilti05 & exp09-cupola & 7:26 & 0.19 \\
hilti06 & exp11-lower-gallery & 2:31 & 0.08 \\
hilti07 & exp15-upper-gallery & 4:20 & 0.13 \\
hilti08 & exp21-outside & 2:32 & 0.14 \\
hilti09 & site1-handheld-1 & 3:24 & 0.17 \\
hilti10 & site1-handheld-2 & 2:47 & 0.15 \\
hilti11 & site1-handheld-3 & 2:50 & 0.15 \\
hilti12 & site1-handheld-4 & 4:55 & 0.28 \\
hilti13 & site1-handheld-5 & 2:39 & 0.14 \\
mars1-1  & HKisland01 & 9:40 & 1.85 \\
mars1-2  & HKisland02 & 5:10 & 1.85 \\
mars1-3  & HKisland03 & 3:46 & 1.85 \\
mars2-1  & HKairport01 & 10:50 & 2.04 \\
mars2-2  & HKairport02 & 5:40& 2.04 \\
mars2-3  & HKairport03 & 4:00 & 2.04 \\
mars3-1  & AMtown01 & 20:00 & 5.11 \\
mars3-2  & AMtown02 & 10:10 & 5.11 \\
mars3-3  & AMtown03 & 7:10 & 5.11 \\
mars4-1  & AMvalley01 & 17:00 & 4.30 \\
mars4-2  & AMvalley02 & 8:45 & 4.30 \\
mars4-3  & AMvalley03 & 6:00 & 4.30 \\
urban1  & UrbanNav-Medium & 13:05 & 3.64 \\
urban2  & UrbanNav-Deep & 25:36 & 4.51 \\
urban3  & UrbanNav-Harsh & 56:07 & 4.86 \\
urban4  & 2019-04-28-20-58-02 & 8:07 & 2.01 \\
urban5  & 2020-03-14-16-45-35 & 5:00 & 1.21 \\
private1 & Jungle-challenge& 2:00 & 0.06 \\
private2 & Campus-elevator & 11:34 & 0.85 \\
\textbf{Total} & & 281:22 & 59.28 \\
\bottomrule
\end{tabular}
}
\end{table}

\bibliography{bare_jrnl}


\end{document}